\RequirePackage{fix-cm}
\documentclass[twocolumn]{svjour3}          
\smartqed  
\usepackage{graphicx}
\usepackage{color}
\usepackage{multirow}
\usepackage{amsmath}
\usepackage{amsmath}
\usepackage{wasysym}

\DeclareMathOperator*{\softmax}{softmax}
\DeclareMathOperator*{\argmax}{argmax}
\DeclareMathOperator*{\argmin}{argmin}

\begin{document}

\title{Teacher-Student Training and Triplet Loss to Reduce the Effect of Drastic Face Occlusion\thanks{This work was supported by a grant of the Romanian Ministry of Education and Research, CNCS - UEFISCDI, project number PN-III-P1-1.1-TE-2019-0235, within PNCDI III. This article has also benefited from the support of the Romanian Young Academy, which is funded by Stiftung Mercator and the Alexander von Humboldt Foundation for the period 2020-2022.}
}
\subtitle{Application to Emotion Recognition, Gender Identification and Age Estimation}


\author{Mariana-Iuliana Georgescu \and
        Georgian-Emilian Du\c{t}\u{a} \and
        Radu Tudor Ionescu}


\institute{M.I. Georgescu \at
              SecurifAI, Bd. Mircea Vod\u{a} 21D, Bucharest, Romania.\\
              Department of Computer Science, University of Bucharest, 14 Academiei, Bucharest, Romania.\\
              \email{georgescu{\_}lily@yahoo.com}           
           \and
              G.E. Du\c{t}\u{a} \at
              SecurifAI, Bd. Mircea Vod\u{a} 21D, Bucharest, Romania.\\
              Department of Computer Science, University of Bucharest, 14 Academiei, Bucharest, Romania.\\
              \email{georgian.duta@securifai.ro}
            \and
              R.T. Ionescu \at
              SecurifAI, Bd. Mircea Vod\u{a}, 21D Bucharest, Romania.\\
              Department of Computer Science and Romanian Young Academy, University of Bucharest, 14 Academiei, Bucharest, Romania.\\
              \email{raducu.ionescu@gmail.com}
}

\date{Received: date / Accepted: date}

\maketitle

\begin{abstract}
We study a series of recognition tasks in two realistic scenarios requiring the analysis of faces under strong occlusion. On the one hand, we aim to recognize facial expressions of people wearing Virtual Reality (VR) headsets. On the other hand, we aim to estimate the age and identify the gender of people wearing surgical masks. For all these tasks, the common ground is that half of the face is occluded. In this challenging setting, we show that convolutional neural networks (CNNs) trained on fully-visible faces exhibit very low performance levels. While fine-tuning the deep learning models on occluded faces is extremely useful, we show that additional performance gains can be obtained by distilling knowledge from models trained on fully-visible faces. To this end, we study two knowledge distillation methods, one based on teacher-student training and one based on triplet loss. Our main contribution consists in a novel approach for knowledge distillation based on triplet loss, which generalizes across models and tasks. Furthermore, we consider combining distilled models learned through conventional teacher-student training or through our novel teacher-student training based on triplet loss. We provide empirical evidence showing that, in most cases, both individual and combined knowledge distillation methods bring statistically significant performance improvements. We conduct experiments with three different neural models (VGG-f, VGG-face, ResNet-50) on various tasks (facial expression recognition, gender recognition, age estimation), showing consistent improvements regardless of the model or task.
\end{abstract}

\section{Introduction}

We aim to study and develop a generic framework suitable for solving various recognition tasks requiring the analysis of faces under strong occlusion. We underline that the studied framework could be useful in several realistic scenarios. In this work, we focus on two specific yet diverse scenarios to demonstrate the practical applicability of our framework. Our first scenario is related to the recognition of facial expressions of people wearing Virtual Reality (VR) headsets. A system able to solve this task with high accuracy provides the means to control and change the VR environment with respect to the user's emotions, in real time. This could be useful for adjusting the level of exposure for VR applications designed for the treatment of various types of phobia. Our second scenario stems from the regulations imposed by many countries around the world to minimize the spread of the SARS-CoV-2 virus, requiring people in public indoor and even outdoor environments to wear surgical masks. In the context of the COVID-19 pandemic, estimating the age and identifying the gender of customers wearing surgical masks is very useful to generate customer demographics for retail stores and supermarkets. Such demographics are necessary for businesses to estimate the impact of advertisement campaigns or to create strategic plans with respect to current trends in customer demand.

In the scenarios enumerated above, the common denominator is the fact that the automated analysis needs to be performed on faces with an occlusion rate of about $50\%$, i.e.~either the upper half or the lower half of the face is occluded. We consider this level of occlusion as \emph{drastic}, given the significant performance damage implied when a deep convolutional neural network (CNN) trained on completely-visible faces is applied on half-visible faces. Perhaps the most natural solution to close the performance gap is to fine-tune the model on occluded faces. 
Yet, we conjecture that the performance gap can be reduced even further by distilling knowledge from a teacher model trained on fully-visible faces into a student model fine-tuned on partially-visible faces. In this work, we study a conventional knowledge distillation technique based on teacher-student training \cite{Hinton-DLRL-2015,Romero-ICLR-2015} as well as a newly developed technique based on triplet loss \cite{Georgescu-ICPR-2020}. In our novel knowledge distillation framework, we formulate the objective such that the model reduces the distance between an anchor embedding, produced by a student CNN that takes occluded faces as input, and a positive embedding (from the same class as the anchor), produced by a teacher CNN trained on fully-visible faces, so that it becomes smaller than the distance between the anchor and a negative embedding (from a different class than the anchor), produced by the student CNN. In addition, we consider combining distilled student models, learned with conventional teacher-student training or triplet loss, into ensemble models.

We conduct experiments on multiple tasks and data sets to demonstrate the generality of using knowledge distillation to reduce the negative effect of face occlusion on accuracy. More precisely, we present facial expression recognition results on AffectNet \cite{Mollahosseini-TAC-2019} and FER+ \cite{Barsoum-ICMI-2016}, gender identification results on UTKFace \cite{Zhang-CVPR-2017} and age estimation results on UTKFace. To simulate occlusions, we replace the pixel values in the upper or lower half of an image with zeros, blacking out the corresponding region as necessary. To demonstrate the generality across models, we consider two VGG architectures, namely VGG-f \cite{Chatfield-BMVC-14} and VGG-face \cite{Parkhi-BMVC-2015}, and one residual architecture, namely ResNet-50 \cite{He-CVPR-2016}. Our empirical results indicate that knowledge distillation yields superior results across the investigated tasks and models.

With respect to our preliminary work \cite{Georgescu-ICPR-2020}, we make the following contributions:
\begin{itemize}
\item We provide a more comprehensive description of the proposed methods.
\item We demonstrate the generality across multiple tasks, adding gender recognition and age estimation as new tasks.
\item We demonstrate the generality across multiple models, considering both residual and VGG-like architectures.
\end{itemize}

The rest of the paper is organized as follows. Section~\ref{sec_related_work} contains an overview of the related scientific articles. The investigated methods are detailed in Section~\ref{sec_method}. The experiments and results are described in Section~\ref{sec_experiments}. Our conclusions and future work directions are provided in Section~\ref{sec_conclusion}.

\section{Related Work}
\label{sec_related_work}

\subsection{Facial Expression Recognition}

Over the past few years, the research efforts on facial expression recognition have concentrated on building and training deep neural networks aimed at obtaining state-of-the-art results~\cite{Ding-FG-2017,Georgescu-Access-2019,Giannopoulos-2018,Hasani-CVPRW-2017,Hosseini-FG-2019,Hua-Access-2019,Kollias-BMVC-2019,Kim-JMUI-2016,Li-MTA-2017,Li-CVPR-2017,Liu-CVPRW-2017,Li-ICPR-2018,Meng-FG-2017,Mollahosseini-CVPRW-2016,She-CVPR-2021,Shi-ACCESS-2021,Siqueira-AAAI-2020,Tang-WREPL-2013,Vo-Access-2020,Wang-CVPR-2020,Wikanningrum-CENIM-2019}. Engineered models based on handcrafted features~\cite{Al-Chanti-VISIGRAPP-2018,Ionescu-WREPL-2013,Shah-PRL-2017,Shao-PRL-2015} have attracted comparably less attention, since such models typically attain less accurate results than deep learning models. In works such as~\cite{Barsoum-ICMI-2016,Guo-Sensors-2020}, the authors adopted VGG-like architectures. Barsoum et al.~\cite{Barsoum-ICMI-2016} proposed a CNN particularly for the FER+ benchmark, formed of $13$ layers (VGG-13). Guo et al.~\cite{Guo-Sensors-2020} concentrated on recognizing facial expressions on mobile devices, designing a light-weight VGG architecture. In order to minimize computational costs, the authors reduced the input size, the number of filters and the number of layers, and replaced the fully-connected layers with global average pooling. Their architecture is formed of $12$ layers divided into $6$ blocks.

Although the majority of works studied facial expression recognition from static images, there is a body of works focusing on video~\cite{Hasani-CVPRW-2017,Kaya-IVC-2017}. Hasani et al.~\cite{Hasani-CVPRW-2017} designed a neural architecture that is formed of 3D convolutional layers followed by a Long Short-Term Memory (LSTM) network, extracting spatial relations within facial images and temporal relations between different frames in a video. 

In a different direction from the aforementioned methods, Meng et al.~\cite{Meng-FG-2017} and Liu et al.~\cite{Liu-CVPRW-2017} presented identity-aware facial expression recognition models. Meng et al.~\cite{Meng-FG-2017} suggested to jointly estimate expression and identity features using a neural architecture consisting of two identical CNN streams, aiming to alleviate inter-subject variations introduced by personal attributes and attain superior facial expression recognition performance. Liu et al.~\cite{Liu-CVPRW-2017} considered deep metric learning, jointly optimizing a deep metric loss and the softmax loss. They obtained an identity-invariant model by using a scheme based on identity-aware hard-negative mining and online positive mining. Li et al.~\cite{Li-CVPR-2017} optimized a CNN architecture using an upgraded back-propagation algorithm that uses a locality preserving loss aiming to pull the neighboring faces from the same class together. Zeng et al.~\cite{Zeng-ECCV-2018} designed a model that addresses the labeling inconsistencies across data sets. In their approach, images are annotated with multiple pseudo-labels, either given by human annotators or predicted by trained models. Then, a facial expression recognition model is trained to fit the latent ground-truth from the inconsistent pseudo-labels. Hua et al.~\cite{Hua-Access-2019} presented a deep learning algorithm consisting of three sub-networks of different depths. Each sub-network is based on an independently-trained CNN.

Different from all the works mentioned so far and many others~\cite{Al-Chanti-VISIGRAPP-2018,Barsoum-ICMI-2016,Ding-FG-2017,Georgescu-Access-2019,Giannopoulos-2018,Hasani-CVPRW-2017,Hosseini-FG-2019,Hua-Access-2019,HU-ACII-2019,Ionescu-WREPL-2013,Kim-JMUI-2016,Kollias-BMVC-2019,Li-MTA-2017,Li-CVPR-2017,Liu-CVPRW-2017,Li-ICPR-2018,Meng-FG-2017,Mollahosseini-CVPRW-2016,Shah-PRL-2017,Shao-PRL-2015,She-CVPR-2021,Shi-ACCESS-2021,Siqueira-AAAI-2020,Tang-WREPL-2013,Vo-Access-2020,Wang-CVPR-2020,Wen-CC-2017,Wikanningrum-CENIM-2019,Yu-ICMI-2015,Zeng-ECCV-2018}, which recognize facial expressions from fully-visible faces, we concentrate on recognizing the emotion by looking only at the lower part of the face. The number of works that focus on facial expression recognition under occlusion is considerably smaller~\cite{Georgescu-ICONIP-2019,Hickson-WACV-2019,Li-ICPR-2018}. Li et al.~\cite{Li-ICPR-2018} applied a model on synthetically occluded images. They designed an end-to-end trainable Patch-Gated CNN to automatically detect the occluded regions and concentrate on the most discriminative non-occluded regions. Unlike Li et al.~\cite{Li-ICPR-2018}, we consider a more difficult scenario in which half of the face is completely occluded. In order to learn effectively in this difficult scenario, we propose to transfer knowledge from teacher models that are trained on fully-visible (non-occluded) faces. 
 
Closer to our method are the approaches designed for the difficult VR setting~\cite{Georgescu-ICONIP-2019,Hickson-WACV-2019,Houshmand-BigMM-2020}, in which a VR headset covers the upper side of the face. Hickson et al.~\cite{Hickson-WACV-2019} proposed a method that analyzes expressions from the eye region. The eye region is captured by an infrared camera mounted inside the VR headset, making the method less generic. Georgescu et al.~\cite{Georgescu-ICONIP-2019} proposed an approach that analyzes the mouth region captured with a standard camera. The same type of approach is adopted by Houshmand et al.~\cite{Houshmand-BigMM-2020}, but the experiments are conducted on a data set that uses face landmarks to apply the occlusion. In this work, we consider the same setting as \cite{Georgescu-ICONIP-2019,Houshmand-BigMM-2020}, investigating the task of facial expression recognition when the upper half of the face is occluded. Unlike these closely related papers \cite{Georgescu-ICONIP-2019,Houshmand-BigMM-2020}, we propose to perform knowledge distillation to produce more accurate CNN models. We study two knowledge distillation techniques to distill information from CNNs trained on fully-visible faces to CNNs trained on occluded faces. To our knowledge, we are the first to apply knowledge distillation in the context of facial expression recognition under strong occlusion. Moreover, we show the generality of the studied methods across different tasks and neural models.

\subsection{Gender Recognition}
 
Gender prediction models are widely used across different domains such as advertising, security and human-computer interaction. Similar to \cite{Abirami-MT-2020,Bhaskar-ICACCI-2015,Georgescu-A-2020,Hacibeyoglu-IJISAE-2018,Ito-APSIPA-2018,Jhang-JIPS-2020,Rafique-ICIC-2019,You-ICDMW-2014}, we focus on gender prediction from facial images. Some of these works \cite{Abirami-MT-2020,Georgescu-A-2020,Hacibeyoglu-IJISAE-2018,Ito-APSIPA-2018,Rafique-ICIC-2019} proposed the use of CNN models to predict the gender.
Abirami et al.~\cite{Abirami-MT-2020} used a CNN model to jointly predict the gender and the age of a person from facial images. Priadana et al.~\cite{Priadana-DASA-2020} proposed the use of CNN models to predict the gender based on profile pictures posted on a social media platform.
Jhang et al.~\cite{Jhang-JIPS-2020} proposed an ensemble of CNNs using a weighted-softmax voting scheme to address the gender prediction task. The weighted-softmax voting scheme is obtained by applying a fully-connected layer on top of the models' predictions. The fully-connected layer is further trained to learn the weights for each model. The weights of the other models are frozen during the additional training process.
Georgescu et al.~\cite{Georgescu-A-2020} proposed a ResNet-50 model based on pyramidal neurons with apical dendrite activations (PyNADA) to address the gender prediction task, among several other tasks. The neurons are equipped with a novel activation function inspired by some recent neuroscience discoveries.

Similarly to the main body of recent works on gender recognition in images, we also employ a CNN model. Different from all the works mentioned so far~\cite{Abirami-MT-2020,Georgescu-A-2020,Hacibeyoglu-IJISAE-2018,Ito-APSIPA-2018,Rafique-ICIC-2019}, which identify the gender in fully-visible faces, we focus on a harder problem, that of predicting the gender in strongly occluded faces. More precisely, we aim to identify the gender of a person wearing a surgical mask, half of the face being occluded. While there are studies on face recognition under occlusion caused by surgical masks \cite{Ding-ACMMM-2020}, or studies on gender recognition under random occlusions \cite{Juefei-CVPRW-2016}, to the best of our knowledge, we are the first to propose knowledge distillation for the gender prediction task under consistently strong occlusion of the lower half of the face.

\subsection{Age Estimation}
 
Age estimation is a very important task across many domains such as advertising, human-computer interaction and security, among many others. There are many works \cite{Geng-TPAMI-2013,Geng-TPAMI-2007,Guo-CVPR-2009a,Guo-CVPR-2009b,Lanitis-TSMC-2004,Nam-Access-2020,Wang-WACV-2015,Xia-TIFS-2020,Zeng-Access-2020} addressing the problem of age estimation of people from facial images. 
Before the era of deep learning, the age estimation task was tackled by extracting handcrafted features from facial images, then applying a regressor or a classifier on top of the extracted features. Guo et al.~\cite{Guo-CVPR-2009a} proposed the use of biologically-inspired features based on Gabor filters to estimate the age of a person from an image. After extracting the bio-inspired features, Guo et al.~\cite{Guo-CVPR-2009a} applied the PCA algorithm to reduce the number of components. In the end, an SVR model is employed on top of the extracted features. In another work based on biologically-inspired features~\cite{Guo-CVPR-2009b}, people were separated based on gender and age groups, proving that the segregation approach improves the performance of the model by a significant margin.

Wang et al.~\cite{Wang-WACV-2015} employed a CNN model to estimate the age of a person, given a facial image. Instead of estimating the age based on the features resulting from the last layer, Wang et al.~\cite{Wang-WACV-2015} extracted features from different layers of the CNN model and concatenated them to obtain an aging pattern. In the end, both SVM and SVR models are applied on top of the aging patterns. Nam et al.~\cite{Nam-Access-2020} also used a CNN model to estimate the age, but before applying the CNN model, the authors employ a Generative Adversarial Network~\cite{Goodfellow-NIPS-2014} to increase the resolution of the image in order to improve the accuracy of the age predictor.

Similarly to the recent age prediction methods~\cite{Nam-Access-2020,Wang-WACV-2015}, we employ a CNN model to estimate the age of people from facial images. However, the main difference is that we are interested in estimating the age under severe occlusion. To the best of our knowledge, we are the first to study the age estimation task on severely occluded images ($50\%$ of the face being occluded).

\subsection{Knowledge Distillation}

Knowledge distillation~\cite{Ba-NIPS-2014,Hinton-DLRL-2015} is a recently studied technique~\cite{Feng-arxiv-2019,Lopez-ICLR-2016,Park-CVPR-2019,Yim-CVPR-2017,You-KDD-2017,Yu-CVPR-2019} that enables the transfer of knowledge between neural networks. Knowledge distillation is a framework that unifies model compression~\cite{Ba-NIPS-2014,Feng-arxiv-2019,Hinton-DLRL-2015} and learning under privileged information~\cite{Lopez-ICLR-2016,vapnik-vashist-nn-2009}, the former one being more popular than the latter. In model compression, knowledge from a large neural model~\cite{Ba-NIPS-2014,Park-CVPR-2019} or an ensemble of large neural networks~\cite{Hinton-DLRL-2015,You-KDD-2017,Yu-CVPR-2019} is distilled into a shallower or thinner neural network that runs efficiently during inference. In learning under privileged information, knowledge from a learning model trained with privileged information (some additional data representation not available at test time) is transferred to another learning model that does not have access to the privileged information. In our paper, we are not interested in compressing neural models, but in learning under privileged information. In particular, we study teacher-student training strategies, in which the teacher neural network can learn from fully-visible faces and the student neural network can learn from occluded faces only. In this context, hidden (occluded) face regions represent the privileged information.

To the best of our knowledge, we are the first to propose the distillation of knowledge using triplet loss. We should underline that there are a few previous works \cite{Feng-arxiv-2019,Park-CVPR-2019,You-KDD-2017,Yu-CVPR-2019} that distilled triplets or the metric space from a teacher network to a student network. Different from these methods, we do not aim to transfer the metric space learned by a teacher network, but to transfer knowledge from the teacher using metric learning, which is fundamentally different.

\section{Methods}
\label{sec_method}

To demonstrate the effectiveness of our approach across various models, we consider three neural network architectures namely, VGG-f~\cite{Chatfield-BMVC-14}, VGG-face~\cite{Parkhi-BMVC-2015} and ResNet-50~\cite{He-CVPR-2016}.
Each of the considered neural architectures is applied on three different tasks that involve drastic face occlusions. Regardless of the task, our training procedure based on knowledge distillation \cite{Ba-NIPS-2014,Georgescu-ICPR-2020,Hinton-DLRL-2015,vapnik-vashist-nn-2009} is organized in a curriculum composed of the following three steps:
\begin{enumerate}
    \item Training a teacher neural model on fully-visible faces.
    \item Fine-tuning the teacher on half-visible faces, resulting in a pre-trained student model. 
    \item Distill knowledge from the teacher model at step 1 into the student model obtained at step 2.
\end{enumerate}

We emphasize that, due to catastrophic forgetting \cite{McCloskey-PLM-1989}, the model fine-tuned at step 2 forgets useful information learned on fully-visible faces. We conjecture that the lost information can be recovered at step 3, by distilling knowledge from the model trained on fully-visible faces. In this context, we refer to the model trained on fully-visible faces as the \emph{teacher} and the model fine-tuned on half-visible faces as the \emph{student}. In our experiments detailed in Section~\ref{sec_experiments}, we present ablation results after pruning steps 2 and 3, one by one.

Our first step is to train the teacher models on fully-visible faces. 
In the second step, the networks are fine-tuned on occluded faces, thus obtaining pre-trained student models ready to undergo knowledge distillation. In the final training step, the students are fine-tuned on occluded faces, simulating either the setting in which people wear a VR headset or the setting in which people wear a surgical mask.


We underline that each student architecture is identical to the corresponding teacher architecture, as our goal is to learn privileged information from the teacher \cite{vapnik-vashist-nn-2009}. Thus, during the training process, we never mix the network architectures, such that when the teacher is a VGG-f network, the corresponding student is also a VGG-f network. The same applies to VGG-face and ResNet-50. 
We do not aim to compress the models, i.e.~to train a shallow and efficient model by distilling a deeper network. Instead, we aim to distill the knowledge from the teacher as privileged information for the student. The privileged information available for the student is the other half of the face (the occluded half) seen only by the teacher. More specifically, for people wearing VR headsets, the upper half of the face represents the privileged information, while for people wearing surgical masks, the privileged information is the lower half of the face.

In the remainder of this section, we describe the two teacher-student training strategies for learning to predict the facial expression, gender or age of people from images with strong face occlusion. The two teacher-student strategies are alternatively employed at step 3 in our training process. As an attempt to increase robustness and stability, we also combine the resulting student models into an ensemble based on meta-learning.

\begin{figure*}[!t]
\centering
\includegraphics[width=0.64\linewidth]{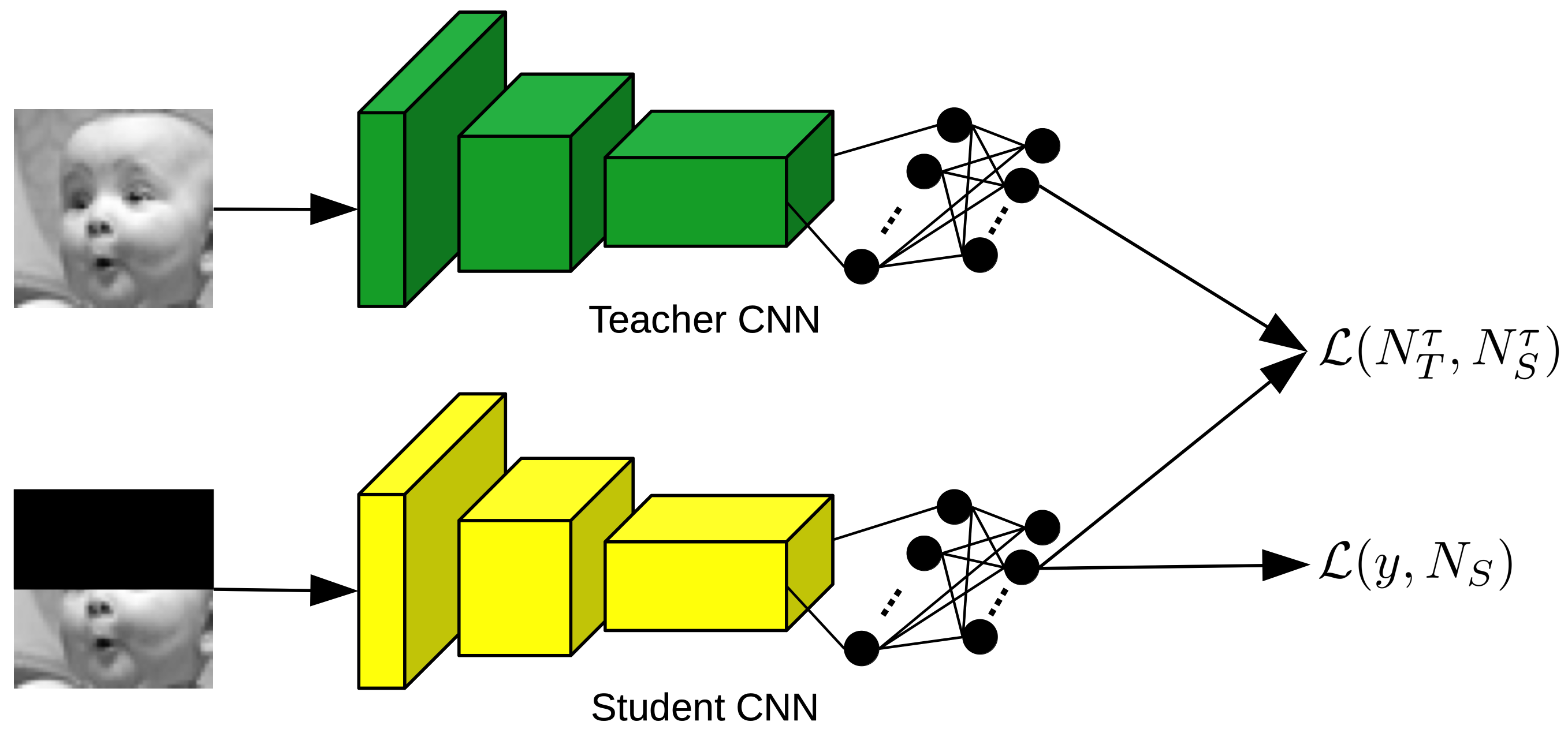}
\caption{The standard teacher-student training pipeline for facial expression recognition on severely occluded faces. The teacher CNN takes as input non-occluded (fully-visible) faces, having access to privileged information. The student CNN takes as input only occluded (lower-half-visible) faces, but learns useful information from the teacher CNN model. The loss functions $\mathcal{L}(y, N_S)$ and $\mathcal{L} (N_{T}^\tau , N_S^{\tau})$ are the terms of the loss defined in Equation~\eqref{eq_loss_KD}. Best viewed in color.}\label{fig_teacher_student}
\end{figure*}

\subsection{Conventional Teacher-Student Training}
\label{sec_train_teacher}

Ba et al.~\cite{Ba-NIPS-2014} proposed a method to compress a deeper model into a shallow one, their purpose being that of training an efficient network to mimic the complex representation of a deeper neural network.
Using a similar idea, Hinton et al.~\cite{Hinton-DLRL-2015} proposed to distill the knowledge of an ensemble of models into a single neural architecture. In their study, the authors demonstrated that, even when an entire class of samples is missing during training, through knowledge distillation, the student is still able to recognize the examples belonging to the missing class due to the knowledge received from the teacher. The knowledge distillation is performed by training the student model using a soft target distribution produced by the teacher model. Besides the knowledge distillation loss, the cross-entropy loss between the ground-truth labels and the student predictions is also added, obtaining the following weighted loss:
\begin{equation}\label{eq_loss_KD}
\mathcal{L}_{KD} (\theta_S) = \lambda \mathcal{L}(N_{T}^\tau, N_S^{\tau}) +  (1 - \lambda) \mathcal{L}(y, N_S),
\end{equation}
where $\theta_S$ are the weights of the student model $S$, $y$ is the ground-truth label, $N_{T}$ is the prediction of the teacher network $T$, $N_S$ is the prediction of the student network $S$ and $\lambda$ is a hyperparameter that controls the importance of each loss function forming the final loss $\mathcal{L}_{KD}$. The first term of $\mathcal{L}_{KD}$ is the cross-entropy with respect to the soft prediction of the teacher and the second term is the cross-entropy with respect to the ground-truth label. $N_{T}^\tau$ and $N_S^{\tau}$ are the softened predictions of the teacher $T$ and the student $S$, respectively, where $\tau > 1$ is a temperature parameter for the softening operation. More precisely, $N_{T}^\tau$ and $N_S^{\tau}$ are derived from the pre-softmax activations $A_T$ and $A_S$ of the teacher network and the student network, as follows:
\begin{equation}
N_{T}^\tau = \softmax \left( \frac{A_T}{\tau} \right), N_{S}^\tau = \softmax \left( \frac{A_S}{\tau} \right).
\end{equation}

Lopez et al.~\cite{Lopez-ICLR-2016} introduced the generalized distillation, where a model can learn from a different teacher model, but also from a different data representation. When training the student model by optimizing the loss defined in Equation~\eqref{eq_loss_KD}, the student can learn privileged information available only to the teacher. In our case, we employ the generalized distillation method to learn a student to recognize facial expressions on strongly occluded faces with privileged information coming from a teacher that has access to fully-visible faces \cite{vapnik-vashist-nn-2009}. The application of the conventional teacher-student framework for facial expression recognition under drastic occlusion is illustrated in Figure~\ref{fig_teacher_student}.

We underline that the aforementioned knowledge distillation methods \cite{Ba-NIPS-2014,Hinton-DLRL-2015,Lopez-ICLR-2016} are suitable for classification problems with multiple classes, in which the softmax output has sufficient components to contain edifying information. As a solution for our regression and binary classification tasks, we distill the knowledge in the penultimate layer of the student. To distill knowledge at any given layer, including the penultimate one, we can employ the approach of Romero et al.~\cite{Romero-ICLR-2015}. 
In order to train a deeper and thinner student than the teacher, Romero et al.~\cite{Romero-ICLR-2015} provided \textit{hints} learned by the teacher network to the student network. A \textit{hint} is the output of a teacher's hidden layer, that is used to guide the corresponding output of a student's hidden layer. The loss function used to guide the training of the student is the $L_1$ distance between the output $H_T$ of a teacher's hidden layer and the output $H_S$ of a student's hidden layer. More precisely, the following weighted loss is proposed by Romero et al.~\cite{Romero-ICLR-2015}:
\begin{equation}\label{eq_loss_hint}
\mathcal{L}_{HT} (\theta_S) = \lambda \lVert H_T - H_S \rVert_1 +  (1 - \lambda) \mathcal{L}(y, N_S),
\end{equation}
where $\theta_S$ are the weights of the student model $S$, $y$ is the ground-truth label, $H_S$ is the output of a student's hidden layer, $H_T$ is the output of the teacher's hint (hidden) layer, $N_S$ is the final output of the student network $S$ and $\lambda$ is a hyperparameter that controls the importance of the two components.

For the age estimation and gender recognition tasks, we employ the knowledge distillation method of Romero et al.~\cite{Romero-ICLR-2015}. We prefer the hint teacher-student paradigm \cite{Romero-ICLR-2015} over the standard teacher-student method \cite{Hinton-DLRL-2015}, due to the low number of available predictions in the final layers for gender recognition (at most two components for male versus female classification) and age estimation (one component for estimating the age on a continuous scale).

 
\subsection{Teacher-Student Training with Triplet Loss}
\label{sec_train_teacher_triplet}

\begin{figure*}[!t]
\centering
\includegraphics[width=1.0\linewidth]{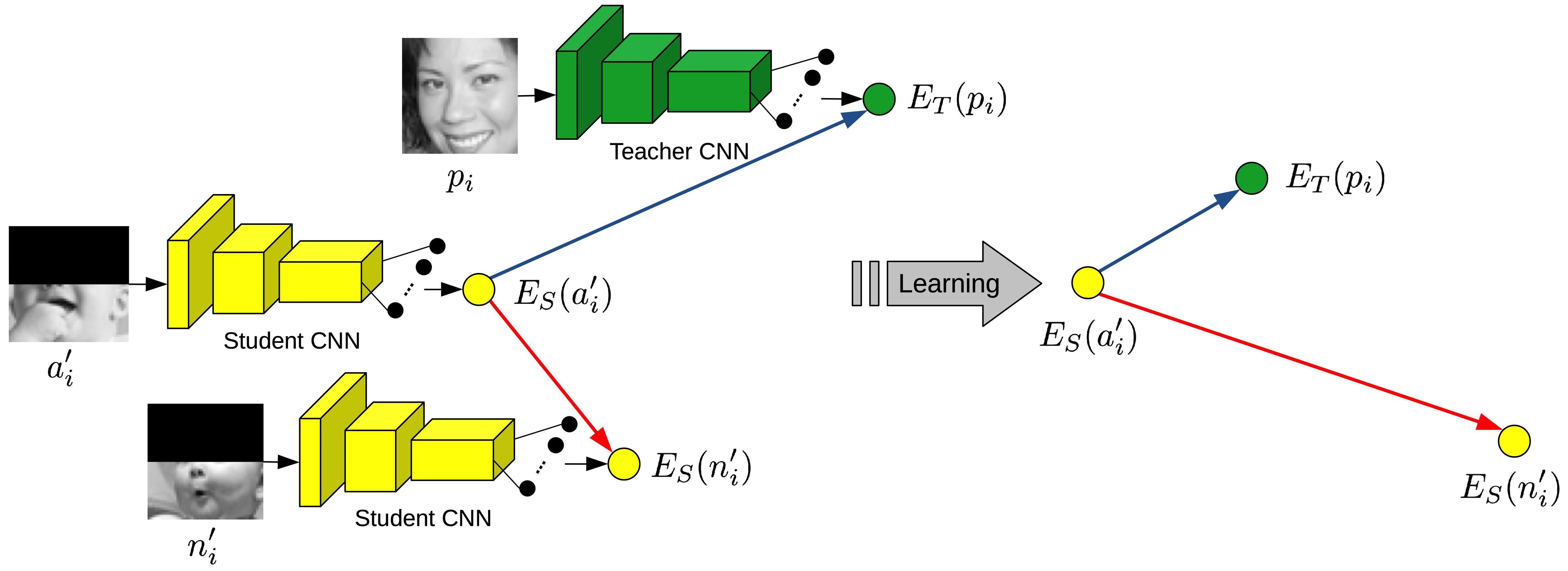}
\caption{The teacher-student training based on triplet loss for facial expression recognition on severely occluded faces. During training, we modify the weights of the student network such that the distance $\lVert E_S(a'_i) - E_T(p_i)\rVert_2^2$ becomes smaller than the distance $\lVert E_S(a_i') - E_S(n'_i)\rVert_2^2$. Best viewed in color.}\label{fig_triplet_loss}
\end{figure*}

When employing the teacher-student paradigm, aside from training the student to predict the correct labels, related techniques add a loss term to minimize the difference between the output of the student and that of the teacher~\cite{Ba-NIPS-2014,Hinton-DLRL-2015,Lopez-ICLR-2016}, or the difference between some intermediate layers~\cite{Romero-ICLR-2015} of the student and teacher networks. To our knowledge, we are the first to perform distillation by adding a triplet loss term. 

In general, triplet loss~\cite{Schroff-CVPR-2015} is employed in neural network training to produce close (similar) embeddings for objects belonging to the same class and distant embeddings when the objects belong to different classes. Our approach is to employ triplet loss on the face embeddings (the activations from the layer immediately before the final  classification or regression layer). We aim to obtain similar face embeddings when the student network and the teacher network take an input image from the same class, and different embeddings otherwise. 

We next present in detail how triplet loss can be applied to train the student network to learn privileged information provided by the teacher network. Throughout the remainder of this section, we use the prime symbol to denote an occluded face. Let $x$ be a fully-visible face and $x'$ be an occluded face. Let $E_T(x)$ and $E_S(x')$ be the face embeddings produced by the teacher network $T$ and the student network $S$, respectively. In order to employ the triplet loss, we need triplets of input images of the form $(a', p, n')$, where $a'$ is an occluded image from a class $k$, $p$ is a fully-visible image from class $k$ (the positive sample) and $n'$ is an occluded image from a different class than $k$ (the negative sample). During training, our goal is to reduce the distance between the anchor embedding $E_S(a')$ and the positive embedding $E_T(p)$ until it becomes smaller than the distance between the anchor embedding $E_S(a')$ and the negative embedding $E_S(n')$. In order to accomplish this goal, we use the following triplet loss function:
\begin{equation}\label{eq_loss_triplet}
\begin{split}
\mathcal{L}_{triplet}(\theta_S) = \sum_{i=1}^{m} \Big[ &\lVert E_S(a'_i) - E_T(p_i)\rVert_2^2 - \\
&- \lVert E_S(a_i') - E_S(n'_i) \rVert_2^2 + \alpha \Big]_+,
\end{split}
\end{equation}
where $\theta_S$ are the learnable parameters of the student network $S$, $m$ is the number of training examples, 
$ [ \cdot ]_+ = max(0, \cdot)$ and $\alpha$ is the margin (minimum distance) between the positive pair of embeddings ($E_S(a'_i), E_T(p_i)$) and the negative pair of embeddings $(E_S(a_i'), E_S(n'_i))$.

Similarly to the standard teacher-student loss expressed in Equation~\eqref{eq_loss_KD}, we want the student network to be able to reproduce the correct labels. Thus, our final loss function becomes:
\begin{equation}\label{eq_loss_KD_triplet}
\mathcal{L}_{KDT} (\theta_S) = (1 - \lambda) \mathcal{L}(y, N_S) + \lambda \mathcal{L}_{triplet}(\theta_S),
\end{equation}
where $\theta_S$ are the learnable parameters of the student network $S$, $\mathcal{L}(y, N_S)$ is the loss function with respect to the ground-truth labels (cross-entropy for classification tasks and mean absolute error for regression tasks), $\mathcal{L}_{triplet}$ is the triplet loss function and $\lambda$ is a hyperparameter that controls the importance of the second objective function with respect to the first one. We illustrate our knowledge distillation paradigm based on triplet loss in Figure \ref{fig_triplet_loss}. We underline that only the weights $\theta_S$ of the student are updated during training, while the embeddings $E_T(p_i)$ are kept unchanged during the whole knowledge distillation process.

Following \cite{Schroff-CVPR-2015}, we propose a fast hard example mining scheme, as described next. In order to speed-up the training process, we generate the triplets offline, at the beginning of each epoch. One by one, each sample $x'$ from the training set is selected as the anchor $a'$. For each anchor $a'$ belonging to the class $k$, we randomly select a subset $S_{pos}$ of fully-visible training faces belonging to the same class. Next, we compute the distance between the embeddings produced by the teacher network for the fully-visible faces in $S_{pos}$ and the embedding produced by the student for the anchor $a'$. The sample which is the farthest from the anchor is selected as the positive example $p$:
\begin{equation}\label{eq_pos_mining}
p = \argmax\limits_j \left\{\lVert E_S(a') - E_T(p_j)\rVert_2^2\right\}, \forall p_j \in S_{pos}.
\end{equation}

In order to select the negative example $n'$, we randomly select a subset $S_{neg}$ of half-visible training faces from a different class than $k$. Then, we compute the distance between the anchor embedding and the embeddings produced by the student model for the half-visible faces in $S_{neg}$. The sample which has the embedding closest to the anchor embedding is selected as the negative example $n'$:
\begin{equation}\label{eq_neg_mining}
n' = \argmin\limits_j \left\{\lVert E_S(a')- E_S(n'_j)\rVert_2^2\right\}, \forall n'_j \in S_{neg}.
\end{equation}

The random subsets $S_{pos}$ and $S_{neg}$ used in our hard example mining scheme contain only a small percentage of the entire training set ($10\%$ for facial expression recognition and $20\%$ for age and gender estimation), speeding up the training time by a large margin. For additional efficiency improvements, we generate the subsets of positive and negative samples only once per epoch. Since age estimation is a regression task, we do not have truly positive or negative examples. Hence, for each anchor example, we consider examples under a difference of 5 years as positive and the other examples as negative. The threshold is set to roughly match the mean absolute error of the teachers on UTKFace.

\section{Experiments}
\label{sec_experiments}

\subsection{Organization}

We hereby present experiments to demonstrate the efficiency of the proposed knowledge distillation methods. We consider three tasks (facial expression recognition, gender recognition, age estimation) and three models (VGG-f, VGG-face, ResNet-50), evaluating all models across all tasks in three scenarios (fully-visible faces, lower-half-visible faces, upper-half-visible faces). For each data set used in our experiments, we also present ablation results, considering as baselines the students trained on occluded faces (without employing any knowledge distillation technique) and the teachers trained on fully-visible faces (without fine-tuning on occluded faces).
Our knowledge distillation frameworks are employed only in the scenarios of interest (facial expression recognition on lower-half-visible faces, gender prediction on upper-half-visible faces, age estimation on upper-half-visible faces).

\subsection{Facial Expression Recognition} 
\subsubsection{Data Sets} 
\noindent
{\bf FER+.}
The FER+ data set~\cite{Barsoum-ICMI-2016} is a curated version of the FER 2013 data set~\cite{Goodfellow-ICONIP-2013}. The latter data set contains images with incorrect labels as well as images not containing faces. Barsoum et al.~\cite{Barsoum-ICMI-2016} cleaned up the FER 2013 data set by relabeling images and by removing those without faces. In the relabeling process, Barsoum et al.~\cite{Barsoum-ICMI-2016} added a new class of emotion, \textit{contempt}, while also keeping the other $7$ classes from FER 2013: \textit{anger}, \textit{disgust}, \textit{fear}, \textit{happiness}, \textit{neutral}, \textit{sadness} and \textit{surprise}. The FER+ data set is composed of 25,045 training images, 3,191 validation images and 3,137 test images. The size of each image is $48 \times 48$ pixels.

\noindent
{\bf AffectNet.}
The AffectNet~\cite{Mollahosseini-TAC-2019} data set is one of the largest data sets for facial expression recognition, containing 287,651 training images and 4,000 validation images with manual annotations. The images from AffectNet have various sizes. Since the test set is not yet publicly available, methods~\cite{Georgescu-Access-2019,Mollahosseini-TAC-2019,She-CVPR-2021,Shi-ACCESS-2021,Vo-Access-2020,Wang-CVPR-2020,Zeng-ECCV-2018} are commonly evaluated on the validation set. The data set contains the same $8$ classes of emotion as FER+. With 500 images per class in the validation set, the class distribution is balanced. In the same time, the training data is unbalanced. As proposed by Mollahosseini et al.~\cite{Mollahosseini-TAC-2019}, we down-sample the training set for classes with more than 15,000 images. This leaves us with a training set of 88,021 images. 

\subsubsection{Data Preprocessing}

In order to train and evaluate the neural models in scenarios with drastic face occlusion, we replace the values of specific pixels with zero to simulate occlusions.

For the setting introduced by Georgescu et al.~\cite{Georgescu-ICONIP-2019}, in which facial expressions are recognized from the lower half of the face, we occlude the entire upper half of the FER+ and AffectNet images. For the setting proposed by Hickson et al.~\cite{Hickson-WACV-2019}, in which facial expressions are recognized from the eye region, we occlude the entire lower half of the FER+ and AffectNet images. 
All images are resized to $224 \times 224$ pixels, irrespective of the data set, in order to be given as input to VGG-f, VGG-face and ResNet-50.

\subsubsection{Evaluation Metrics}

Our metric for evaluating the classification models for facial expression recognition is the \textit{accuracy} between the ground-truth labels and the predicted labels. Due to the fact that FER+ is highly imbalanced, we also report the \textit{weighted accuracy} for this data set.  

\subsubsection{Implementation Details} 

We emphasize that all the hyperparameters specified below are tuned on the FER+ validation set. Since the AffectNet validation set is used for the final evaluation, on AffectNet, we use the hyperparameter settings found optimal on the FER+ validation set. The VGG-f and VGG-face models are trained with stochastic gradient descent with momentum. We set the momentum rate to $0.9$. The VGG-face model is trained on mini-batches of $64$ images, while the VGG-f model is trained on mini-batches of $512$ images, since the latter model has a lower memory footprint. We use the same mini-batch sizes in all training stages.
The ResNet-50 model is trained using the Adam optimizer \cite{Kingma-ICLR-2015} on mini-batches of $16$ images.

\noindent
{\bf Preliminary training of teachers and students.}
For the preliminary fine-tuning of the teacher and the student models, we use the MatConvNet~\cite{matconvnet} library. The teacher VGG-face is fine-tuned on facial expression recognition from fully-visible faces for a total of $50$ epochs. The teacher VGG-f is fine-tuned for $800$ epochs. The student VGG-face is fine-tuned on facial expression recognition from occluded faces for $40$ epochs. Similarly, the student VGG-f is fine-tuned on occluded faces for $80$ epochs. Further details about training these VGG-face and VGG-f baselines on fully-visible or occluded faces are provided in~\cite{Georgescu-ICONIP-2019}. The teacher ResNet-50 is fine-tuned for $75$ epochs using a learning rate of $10^{-4}$. Similarly, the student ResNet-50 is fine-tuned on lower-half-visible faces for $75$ epochs with a learning rate set to $10^{-4}$. 

\noindent
{\bf Standard teacher-student training.}
For the conventional teacher-student strategy, the student VGG-face is trained for $50$ epochs starting with a learning rate of $10^{-4}$, decreasing it when the validation error does not improve for $10$ consecutive epochs. By the end of the training process, the learning rate for the student VGG-face drops to $10^{-5}$. In a similar manner, the student VGG-f is trained for $200$ epochs starting with a learning rate of $10^{-3}$, decreasing it when the validation error does not improve for $10$ consecutive epochs. By the end of the training process, the learning rate for the student VGG-f drops to $10^{-4}$. The student ResNet-50 is trained for $10$ epochs starting with a learning rate of $10^{-5}$. Hinton et al.~\cite{Hinton-DLRL-2015} suggested to use a lower weight on the second objective function defined in Equation~\eqref{eq_loss_KD}. Therefore, we set $\lambda$ to a value of $0.9$ for ResNet-50 and a value of $0.7$ for both VGG-f and VGG-face models. The parameter $\lambda$ is validated on the FER+ validation set.

\noindent
{\bf Teacher-student training with triplet loss.}
To implement the teacher-student training based on triplet loss, we switch to TensorFlow \cite{Abadi-OSDI-2016}, exporting the {VGG-f} and VGG-face models from MatConvNet. We train the student VGG-face for $10$ epochs using a learning rate of $10^{-6}$. In a similar fashion, we train the VGG-f and ResNet-50 students for $10$ epochs using a learning rate of $10^{-5}$. The parameter $\lambda$ in Equation~\eqref{eq_loss_KD_triplet} is set to a value of $0.5$ for the VGG-f and VGG-face models, and a value of $0.1$ for the ResNet-50 model. The value of the margin $\alpha$ from Equation~\eqref{eq_loss_triplet} is chosen based on the performance measured on the validation sets, considering values in the interval $[0,0.5]$ at a step of $0.1$.

\noindent
{\bf Combining distilled embeddings.}
After training the student models using the two teacher-student strategies independently, we concatenate the corresponding face embeddings into a single embedding vector. The concatenated embeddings are provided as input to a Support Vector Machines (SVM) model~\cite{cortes-vapnik-ml-1995}. The regularization parameter C of the resulting SVM models is chosen according to the performance on the validation sets, considering values between $10^{-1}$ and $10^3$, at a step of $10^1$. We use the SVM implementation from Scikit-learn~\cite{Pedregosa-JMLR-2011}.

\begin{table*}[!t]
\caption{Accuracy rates of various models on AffectNet~\cite{Mollahosseini-TAC-2019} and FER+~\cite{Barsoum-ICMI-2016}, for fully-visible faces (denoted by {\Circle}), lower-half-visible faces (denoted by {\protect\rotatebox[origin=c]{90}{\RIGHTcircle}}) and upper-half-visible faces (denoted by {\protect\rotatebox[origin=c]{90}{\LEFTcircle}}). The VGG-f, VGG-face and ResNet-50 models based on our teacher-student (T-S) training strategies are compared with state-of-the-art methods~\cite{Barsoum-ICMI-2016,Georgescu-Access-2019,Guo-Sensors-2020,Farzaneh-WACV-2021,Ionescu-WREPL-2013,Kollias-BMVC-2019,Mollahosseini-TAC-2019,She-CVPR-2021,Shi-ACCESS-2021,Siqueira-AAAI-2020,Vo-Access-2020,Wang-CVPR-2020} tested on fully-visible faces and with methods~\cite{Georgescu-ICONIP-2019,Hickson-WACV-2019} designed for the VR setting (tested on occluded faces). The test results of our student networks that are significantly better than the stronger baseline~\cite{Georgescu-ICONIP-2019}, according to a paired McNemar's test \cite{Dietterich-NC-1998}, are marked with $\dagger$ for a significance level of $0.05$.}\label{tab_results}
\setlength\tabcolsep{5.5pt} %
\begin{center}
\begin{tabular}{|l|c|c|c|c|c|} %
\hline 
\multirow{2}{*}{\bf Model}    & \multirow{2}{*}{\bf Train faces}    &  \multirow{2}{*}{\bf Test faces}     &  \multirow{2}{*}{\bf AffectNet}     &  \multicolumn{2}{|c|}{\bf FER+}\\
\cline{5-6} 
             &                    &                    &                   & \bf Accuracy   & \bf Weighted accuracy \\
\hline
\hline
VGG-13~\cite{Barsoum-ICMI-2016}                & {\Circle}  & {\Circle} & -        & $84.99\%$  & - \\
DACL~\cite{Farzaneh-WACV-2021}                 & {\Circle}  & {\Circle}  & $65.20\%$         & - & - \\
CNNs+BOVW+LC~\cite{Georgescu-Access-2019}  & {\Circle}  & {\Circle}  & $59.58\%$         & $87.76\%$ & - \\
VGG-12~\cite{Guo-Sensors-2020}                 & {\Circle}  & {\Circle}  & $58.50\%$         & - & - \\
Bag-of-visual-words~\cite{Ionescu-WREPL-2013}  & {\Circle}  & {\Circle} & $48.30\%$ & $80.65\%$ & - \\
MT-VGG~\cite{Kollias-BMVC-2019}                & {\Circle}  & {\Circle} & $54.00\%$         & - & - \\
AlexNet~\cite{Mollahosseini-TAC-2019}          & {\Circle}  & {\Circle} & $58.00\%$ & -    & - \\    
Res-50IBN~\cite{She-CVPR-2021}                 & {\Circle}  & {\Circle}  & $63.11\%$         & $89.51\%$ & - \\
MBCC-CNN~\cite{Shi-ACCESS-2021}                & {\Circle}  & {\Circle}  &    -              & $88.10\%$  & - \\
ESR-9~\cite{Siqueira-AAAI-2020}                & {\Circle}  & {\Circle}  & $59.30\%$         & $87.15\%$ & - \\ 
SCN~\cite{Wang-CVPR-2020}                      & {\Circle}  & {\Circle}  & $60.23\%$         & $89.35\%$ & - \\
PSR~\cite{Vo-Access-2020}                      & {\Circle}  & {\Circle}  & $60.68\%$         & $89.75\%$ & - \\
 
\hline
Teacher VGG-f                                  & {\Circle}  & {\Circle}    & $57.37\%$         & $85.05\%$  & $59.71$\%\\
Teacher VGG-face                               & {\Circle}  & {\Circle}    & $59.03\%$         & $84.79\%$  & $66.15\%$ \\ 
Teacher ResNet-50                              & {\Circle}  & {\Circle}    & $56.07\%$         & $85.91\%$  & $65.67\%$ \\ 
\hline
Teacher VGG-f         & {\Circle}    & {\rotatebox[origin=c]{90}{\RIGHTcircle}}   & $41.58\%$         & $70.00\%$ & $43.24\%$ \\
Teacher VGG-face      & {\Circle}    & {\rotatebox[origin=c]{90}{\RIGHTcircle}}   & $37.70\%$         & $68.89\%$ & $39.69\%$ \\
Teacher ResNet-50     & {\Circle}    & {\rotatebox[origin=c]{90}{\RIGHTcircle}}   & $40.50\%$         & $70.89\%$ & $44.56\%$ \\
\hline
Teacher VGG-f         & {\Circle}    & {\rotatebox[origin=c]{90}{\LEFTcircle}}   & $26.85\%$         & $40.07\%$ & $32.82\%$ \\
Teacher VGG-face      & {\Circle}    & {\rotatebox[origin=c]{90}{\LEFTcircle}}   & $31.23\%$         & $48.29\%$ & $37.36\%$ \\
Teacher ResNet-50     & {\Circle}    & {\rotatebox[origin=c]{90}{\LEFTcircle}}   & $24.12\%$         & $44.21\%$ & $30.01\%$ \\
\hline
VGG-f~\cite{Georgescu-ICONIP-2019}       & {\rotatebox[origin=c]{90}{\RIGHTcircle}}       & {\rotatebox[origin=c]{90}{\RIGHTcircle}}  & $47.58\%$         & $78.23\%$  & $50.52\%$\\
VGG-face~\cite{Georgescu-ICONIP-2019}    & {\rotatebox[origin=c]{90}{\RIGHTcircle}}       & {\rotatebox[origin=c]{90}{\RIGHTcircle}}  & $49.23\%$         & $82.28\%$  & $58.69\%$\\
ResNet-50    & {\rotatebox[origin=c]{90}{\RIGHTcircle}}       & {\rotatebox[origin=c]{90}{\RIGHTcircle}}  & $45.90\%$         & $81.79\%$  & $60.57\%$\\
\hline
VGG-f~\cite{Hickson-WACV-2019}           & {\rotatebox[origin=c]{90}{\LEFTcircle}}       & {\rotatebox[origin=c]{90}{\LEFTcircle}}  & $42.45\%$         & $66.18\%$  & $44.66\%$ \\
VGG-face~\cite{Hickson-WACV-2019}        & {\rotatebox[origin=c]{90}{\LEFTcircle}}        & {\rotatebox[origin=c]{90}{\LEFTcircle}}  & $43.18\%$         & $70.19\%$ & $48.83\%$\\
ResNet-50        & {\rotatebox[origin=c]{90}{\LEFTcircle}}        & {\rotatebox[origin=c]{90}{\LEFTcircle}}  & $43.37\%$         & $72.26\%$ & $54.62\%$\\
\hline
VGG-f (standard T-S)                     & {\Circle} + {\rotatebox[origin=c]{90}{\RIGHTcircle}}     & {\rotatebox[origin=c]{90}{\RIGHTcircle}}  & $48.75\%^\dagger$         & $80.17\%^\dagger$  & $53.00\%^\dagger$ \\
VGG-face (standard T-S)                  & {\Circle} + {\rotatebox[origin=c]{90}{\RIGHTcircle}}    & {\rotatebox[origin=c]{90}{\RIGHTcircle}}  & $49.75\%$         & $82.37\%$  &  $59.46\%$\\
ResNet-50 (standard T-S)   & {\Circle} + {\rotatebox[origin=c]{90}{\RIGHTcircle}}    & {\rotatebox[origin=c]{90}{\RIGHTcircle}}  & $46.95\%^\dagger$         & $82.37\%^\dagger$  &  $59.19\%^\dagger$\\
\hline
VGG-f (triplet loss T-S)                 & {\Circle} + {\rotatebox[origin=c]{90}{\RIGHTcircle}}            & {\rotatebox[origin=c]{90}{\RIGHTcircle}}  & $48.13\%$         & $80.05\%^\dagger$ & $52.87\%^\dagger$\\
VGG-face (triplet loss T-S)              & {\Circle} + {\rotatebox[origin=c]{90}{\RIGHTcircle}}           & {\rotatebox[origin=c]{90}{\RIGHTcircle}}  & $49.71\%$         & $82.57\%$  & $59.12\%$\\
ResNet-50 (triplet loss T-S)              & {\Circle} + {\rotatebox[origin=c]{90}{\RIGHTcircle}}           & {\rotatebox[origin=c]{90}{\RIGHTcircle}}  & $46.17\%$         & $81.28\%$  & $60.93\%$\\
\hline
VGG-f (triplet loss + standard T-S)      & {\Circle} + {\rotatebox[origin=c]{90}{\RIGHTcircle}}    & {\rotatebox[origin=c]{90}{\RIGHTcircle}}  & $48.70\%^\dagger$         & $81.09\%^\dagger$ & $58.90\%^\dagger$\\
VGG-face (triplet loss + standard T-S)   & {\Circle} + {\rotatebox[origin=c]{90}{\RIGHTcircle}}      & {\rotatebox[origin=c]{90}{\RIGHTcircle}}  &  $50.09\%^\dagger$        & $82.75\%^\dagger$ & $61.23\%^\dagger$\\
ResNet-50 (triplet loss + standard T-S)   & {\Circle} + {\rotatebox[origin=c]{90}{\RIGHTcircle}}      & {\rotatebox[origin=c]{90}{\RIGHTcircle}}  &  $47.00\%^\dagger$        & $82.37\%^\dagger$ & $59.30\%^\dagger$\\
\hline
\end{tabular}
\end{center}
\end{table*}

\subsubsection{Baselines}
As baselines for facial expression recognition, we consider two state-of-the-art methods~\cite{Georgescu-ICONIP-2019,Hickson-WACV-2019} designed for facial expression recognition in the VR setting. The key contribution of these methods resides in the region they use to extract features, the lower half of the face (mouth region)~\cite{Georgescu-ICONIP-2019} or the upper half of the face (eye region)~\cite{Hickson-WACV-2019}. In order to conduct a fair comparison, we use the same neural architectures for both baselines and our approach.

As reference, we include some results on FER+ and AffectNet from the recent literature~\cite{Barsoum-ICMI-2016,Farzaneh-WACV-2021,Georgescu-Access-2019,Guo-Sensors-2020,Ionescu-WREPL-2013,Kollias-BMVC-2019,Mollahosseini-TAC-2019,She-CVPR-2021,Shi-ACCESS-2021,Siqueira-AAAI-2020,Vo-Access-2020,Wang-CVPR-2020}. We underline that these state-of-the-art methods are trained and tested on fully-visible faces. Hence, the comparison to our approach or other approaches applied on occluded faces~\cite{Georgescu-ICONIP-2019,Hickson-WACV-2019} is unfair, but we included it as a relevant indicator of the upper bound for the models applied on occluded faces. We also underline that the included state-of-the-art methods~\cite{She-CVPR-2021,Shi-ACCESS-2021,Siqueira-AAAI-2020,Vo-Access-2020,Wang-CVPR-2020} are not based on standard modeling choices, using sophisticated loss functions, label smoothing, ensembles of multiple neural architectures or more of the above. Our approach is closer to the methods trained under more simple settings~\cite{Barsoum-ICMI-2016,Guo-Sensors-2020,Mollahosseini-TAC-2019}, since our teachers and students are trained using fairly well-known loss functions and the evaluation is performed using only one model.


\subsubsection{Results} 

In Table~\ref{tab_results}, we present the empirical results obtained on AffectNet~\cite{Mollahosseini-TAC-2019} and FER+~\cite{Barsoum-ICMI-2016} by the VGG-f, VGG-face and ResNet-50 models based on our teacher-student training strategies in comparison with the results obtained by the state-of-the-art methods~\cite{Barsoum-ICMI-2016,Guo-Sensors-2020,Farzaneh-WACV-2021,Ionescu-WREPL-2013,Kollias-BMVC-2019,Mollahosseini-TAC-2019,She-CVPR-2021,Shi-ACCESS-2021,Siqueira-AAAI-2020,Vo-Access-2020,Wang-CVPR-2020} tested on fully-visible faces and by the methods~\cite{Georgescu-ICONIP-2019,Hickson-WACV-2019} designed for the VR setting (tested on occluded faces). 

\noindent
{\bf Comparison with the state-of-the-art.}
First of all, we note that it is natural for the state-of-the-art methods~\cite{Barsoum-ICMI-2016,Guo-Sensors-2020,Farzaneh-WACV-2021,Ionescu-WREPL-2013,Kollias-BMVC-2019,Mollahosseini-TAC-2019,She-CVPR-2021,Shi-ACCESS-2021,Siqueira-AAAI-2020,Vo-Access-2020,Wang-CVPR-2020} to achieve better accuracy rates (on fully-visible faces) than our approach or the other approaches applied on occluded faces~\cite{Georgescu-ICONIP-2019,Hickson-WACV-2019}. One exceptional case is represented by the VGG-face model of Georgescu et al.~\cite{Georgescu-ICONIP-2019} and our student VGG-face model, as both of them surpass the bag-of-visual-words model~\cite{Ionescu-WREPL-2013} on both data sets. Another exception of the above observation is represented by the fine-tuned or distilled ResNet-50 students, both surpassing the bag-of-visual-words on FER+. 

\noindent
{\bf Comparison between lower-half and upper-half visible faces.}
With respect to the baselines~\cite{Georgescu-ICONIP-2019,Hickson-WACV-2019} designed for the VR setting, all our teacher-student training strategies provide superior results. We observe that the accuracy rates of Hickson et al.~\cite{Hickson-WACV-2019} are considerably lower than the accuracy rates of Georgescu et al.~\cite{Georgescu-ICONIP-2019} (differences are between $5\%$ and $12\%$), although the neural models have identical architectures. We hypothesize that this difference is caused by the fact that it is significantly harder to recognize facial expressions from the eye region (denoted by {\rotatebox[origin=c]{90}{\LEFTcircle}}) than from the mouth region (denoted by {\rotatebox[origin=c]{90}{\RIGHTcircle}}). To test this hypothesis, we evaluate the teachers (VGG-f,  VGG-face and ResNet-50) on upper-half-visible and lower-half-visible faces. We observe even larger differences between the results on upper-half-visible faces (accuracy rates are between $24\%$ and $49\%$) and the results on lower-half-visible faces (accuracy rates are between $37\%$ and $71\%$), confirming our hypothesis. We also underline that the results attained by the teacher models on occluded faces are considerably lower than the results of the baselines~\cite{Georgescu-ICONIP-2019,Hickson-WACV-2019} designed for the VR setting, although the teacher models attain  results close to the state-of-the-art methods~\cite{Barsoum-ICMI-2016,Guo-Sensors-2020,Farzaneh-WACV-2021,Ionescu-WREPL-2013,Kollias-BMVC-2019,Mollahosseini-TAC-2019,She-CVPR-2021,Shi-ACCESS-2021,Siqueira-AAAI-2020,Vo-Access-2020,Wang-CVPR-2020}
when testing is performed on fully-visible faces. This indicates that CNN models trained on fully-visible faces are not particularly suitable to handle severe facial occlusions, justifying the need for training on occluded faces.

\begin{figure*}
\centering
  \includegraphics[width=1.0\textwidth]{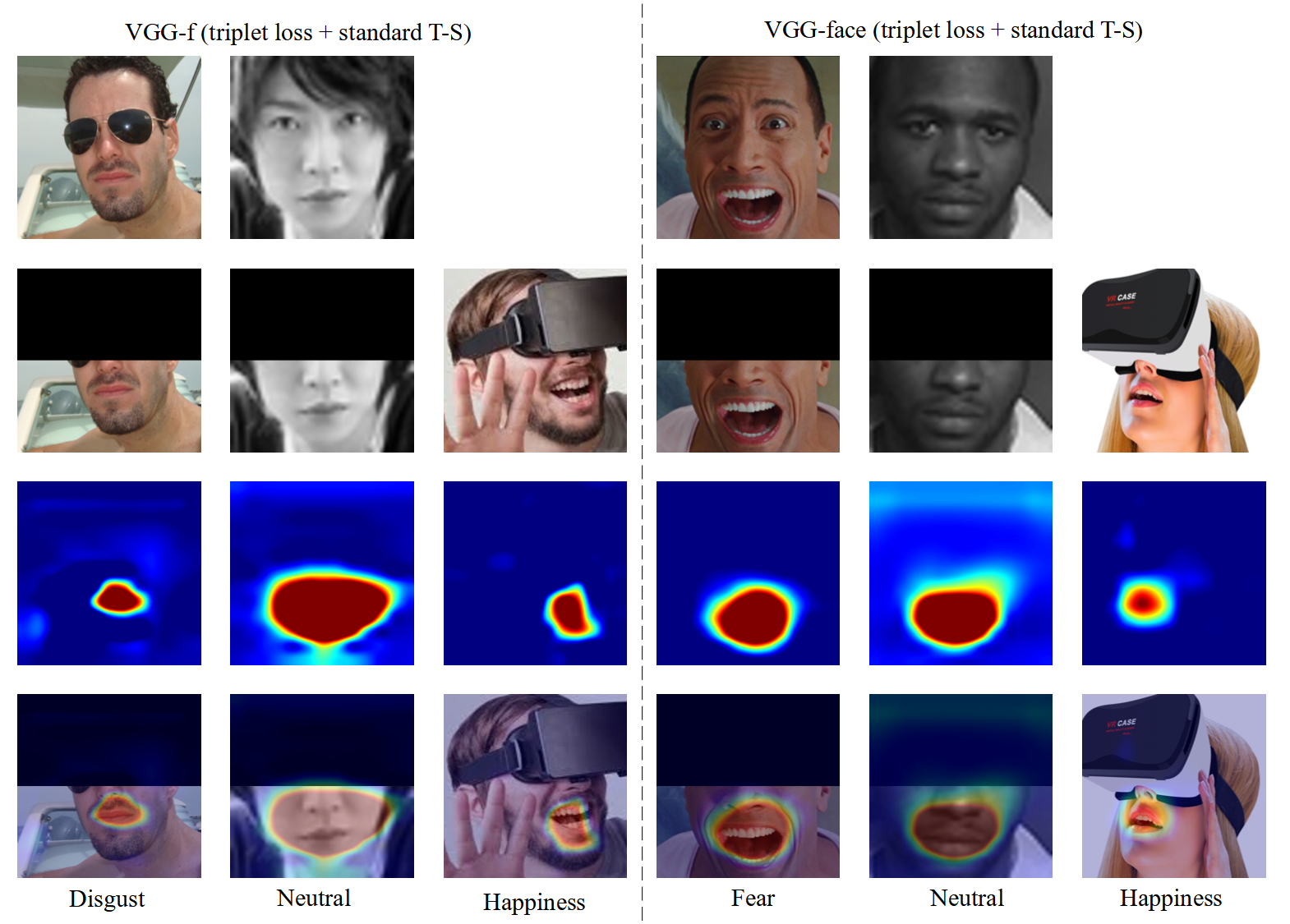}  %
\caption{Fully-visible images ({\Circle}) on top row, lower-half-visible faces ({\protect\rotatebox[origin=c]{90}{\RIGHTcircle}}) on second row, Grad-CAM~\cite{Selvaraju-ICCV-2017} explanation masks on third row and lower-half-visible faces with superimposed Grad-CAM masks on bottom row. The predicted labels provided by the distilled VGG-face (left-hand side) or VGG-f (right-hand side) models are also provided at the bottom. The first two examples from each side are selected from AffectNet~\cite{Mollahosseini-TAC-2019} and FER+~\cite{Barsoum-ICMI-2016}, respectively. The third example from each side is a person wearing an actual VR headset. Best viewed in color.}
\label{fig:fer}       
\end{figure*}

\noindent
{\bf Comparison with closely related methods.}
The results presented in the last nine rows of Table~\ref{tab_results} indicate that the teacher-student learning strategies provide very good results on lower-half-visible faces, surpassing the other methods~\cite{Georgescu-ICONIP-2019,Hickson-WACV-2019} evaluated on occluded faces. We believe that the accuracy gains are due to the teacher neural networks that are trained on fully-visible images, which bring additional (privileged) information from the (unseen) upper half of the training faces. Our teacher-student training strategy based on triplet loss provides results that are comparable to the standard teacher-student training strategy. We also achieve additional performance gains when the two teacher-student strategies are combined through embedding concatenation. Our final models based on the concatenated distilled embeddings attain results that are close to some state-of-the-art methods~\cite{Barsoum-ICMI-2016,Guo-Sensors-2020,Ionescu-WREPL-2013,Kollias-BMVC-2019,Mollahosseini-TAC-2019}. For example, our VGG-face with triplet loss and standard teacher-student training yields an accuracy rate of $82.75\%$ on FER+, which is $2.24\%$ under the state-of-the-art VGG-13~\cite{Barsoum-ICMI-2016}. We thus conclude that our models can recognize facial expressions with sufficient reliability, despite being tested on faces that are severely occluded (the entire upper half is occluded).

\noindent
{\bf Statistical significance testing.}
We also performed statistical significance testing to compare our models (VGG-f, VGG-face and ResNet-50) based on teacher-student training with the models of Georgescu et al.~\cite{Georgescu-ICONIP-2019}, which are equivalent with our students before undergoing distillation. Notably, the combined teacher-student strategies provide significant improvements for the VGG-f, VGG-face and ResNet-50 models on both data sets, with a significance level of $0.05$.

\noindent
{\bf Grad-CAM visualizations.}
In order to better understand how our models make decisions, we used the Grad-CAM~\cite{Selvaraju-ICCV-2017} approach to provide visual explanations for some image examples illustrated in Figure~\ref{fig:fer}. First, we notice that we, as humans, are still able to recognize the facial expressions in the presented examples, even if the upper half of each face depicted in the second row of Figure~\ref{fig:fer} is occluded. We observe that the neural architectures turn their attention on the lower part of the face, particularly on the mouth region. This indicates that our neural networks can properly handle situations in which people wear VR headsets. We note that the predicted labels for the first five samples presented in Figure~\ref{fig:fer} are correct.

\subsection{Age and Gender Estimation} 
\subsubsection{Data Sets}   
The UTKFace~\cite{Zhang-CVPR-2017} data set contains images with faces of people of various age, gender and ethnicity. The data set consists of 23,689 images. We randomly divide the data set obtaining  14,213 ($60\%$) images for training, 4,738 ($20\%$) images for validation and 4,738 ($20\%$) images for testing. The size of each image is $200 \times 200$ pixels. We perform two types of experiments on UTKFace: gender recognition and age estimation.
 
\subsubsection{Data Preprocessing} 
We adopt a similar technique as in the VR setting to simulate occlusions caused by surgical masks, thus replacing the values of pixels in the lower half of each face in the UTKFace data set with zero. The images from the data set are resized to $224 \times 224$ pixels to correspond to the input size of the convolutional networks.

\subsubsection{Evaluation Metrics}

Our metric for evaluating the gender prediction models is the standard \textit{classification accuracy}. To evaluate the regression models for age estimation, we consider the \textit{mean absolute error} (MAE) between the predicted age and the target age of each test sample. 

\subsubsection{Implementation Details} 

The networks are trained using the same optimizer and batch size as in the facial expression recognition experiments. We tune all other hyperparameters on the UTKFace validation set.

\noindent
{\bf Preliminary training of teachers and students.}  
The teacher and student VGG-f models are fine-tuned for $200$ epochs starting with a learning rate of $10^{-4}$. We note that there is a teacher and a student for each of the two tasks (gender recognition and age estimation). Both teacher and student VGG-face networks are fine-tuned for $250$ epochs with the learning rate set to $10^{-4}$. The teacher and student ResNet-50 models are trained from scratch to predict the gender or the age of people in images, for $100$ epochs. The learning rate for all teacher and student ResNet-50 models is set to $10^{-4}$.
  
\noindent
{\bf Standard teacher-student training.}  
In order to apply the standard teacher-student strategy to estimate the age and the gender from an image containing a face, the VGG-f students are each trained for $20$ epochs starting with a learning rate of $10^{-6}$. We set the parameter $\lambda$ in Equation~\eqref{eq_loss_hint} to a value of $0.3$ for gender prediction and a value of $0.5$ for the age estimation task. 
The VGG-face students are trained for $40$ epochs, setting the learning rate to a value of $10^{-5}$ for the age estimation task and a value of $10^{-6}$ for the gender prediction task. The parameter $\lambda$ in Equation~\eqref{eq_loss_hint} is set to $0.3$ for both tasks.
The ResNet-50 students are each trained for a number of $40$ epochs to estimate the age or the gender of a person, respectively. The starting learning rate is $10^{-5}$ and we opted to decrease it when the validation error does not improve for $10$ consecutive epochs. The parameter $\lambda$ in Equation~\eqref{eq_loss_hint} is set to $0.1$ for both tasks, based on the performance observed on the UTKFace validation set.

\noindent
{\bf Teacher-student training with triplet loss.} 
The VGG-f model is trained for $30$ epochs for the gender prediction task, starting with a learning rate of $10^{-5}$. For the age estimation task, the model is trained for $40$ epochs and the learning rate is set to $10^{-6}$. 
The VGG-face model is trained for $40$ epochs with the learning rate set to $10^{-6}$ for both tasks. The parameter $\lambda$ in Equation~\eqref{eq_loss_KD_triplet} is set to $0.7$ for both tasks and both VGG networks.
Each of the two student ResNet-50 models is trained for $40$ epochs with a learning rate of $10^{-5}$. The value of the parameter $\lambda$ is $0.7$ for the age estimation task and $0.4$ for the gender prediction task. The value of the margin $\alpha$ in Equation~\eqref{eq_loss_triplet} is tuned on the validation set, considering values in the range $[0,0.5]$ at a step of $0.1$.
 
\noindent
{\bf Combining distilled embeddings.}  
We concatenate the embeddings obtained from our two teacher-student training strategies and provide them as input either to an SVM model~\cite{cortes-vapnik-ml-1995} for the classification task (gender recognition) or to an $\epsilon$-Support Vector Regression (SVR) for the regression task (age estimation). The regularization parameter C of these models is chosen according to the performance on the validation sets, considering values between $10^{-1}$ and $10^3$, at a step of $10^1$.
We keep the default value for the parameter $\epsilon$ of the SVR, that is $\epsilon=0$. As for the SVM, we employ the SVR implementation from Scikit-learn~\cite{Pedregosa-JMLR-2011}.

\subsubsection{Baselines}

As reference, we include the state-of-the-art results of the ResNet-50 based on pyramidal neurons with apical dendrite activations (PyNADA) reported in~\cite{Georgescu-A-2020}, although the results are not directly comparable due to the different splits applied on UTKFace. As baselines, we also include the student trained on lower-half-visible faces and the teacher trained on fully-visible faces.

\begin{table*} 
\caption{Accuracy rates for gender prediction on UTKFace~\cite{Zhang-CVPR-2017}, for fully-visible faces (denoted by {\Circle}), lower-half-visible faces (denoted by {\protect\rotatebox[origin=c]{90}{\RIGHTcircle}}) and upper-half-visible faces (denoted by {\protect\rotatebox[origin=c]{90}{\LEFTcircle}}). A state-of-the-art model \cite{Georgescu-A-2020} is included as reference. The results of distilled models that are significantly better than the student trained on upper-half-visible faces, according to a paired McNemar's test \cite{Dietterich-NC-1998} at a significance level of $0.05$, are marked with $\dagger$.}
\label{tab:gender_resnet_50} 
\begin{center}
\begin{tabular}{|l|c|c|c|} 
\hline
\bf Method    & \bf Train faces       & \bf Test faces                             &  \bf Accuracy \\
\hline
\hline
ResNet-50+PyNADA \cite{Georgescu-A-2020}    &  {\Circle}     &  {\Circle}                                 & $90.80\%$ \\ 
\hline 
Teacher VGG-f     &  {\Circle}   &  {\Circle}        & $92.78\%$ \\ 
Teacher VGG-face  &  {\Circle}   &  {\Circle}        & $92.20\%$ \\ 
Teacher ResNet-50 &  {\Circle}   &  {\Circle}        & $90.88\%$ \\ 
\hline 
Teacher VGG-f     &  {\Circle}   &  {\rotatebox[origin=c]{90}{\RIGHTcircle}}  & $78.13\%$ \\ 
Teacher VGG-face  &  {\Circle}   &  {\rotatebox[origin=c]{90}{\RIGHTcircle}}  & $73.05\%$ \\ 
Teacher ResNet-50 &  {\Circle}   &  {\rotatebox[origin=c]{90}{\RIGHTcircle}}  & $72.69\%$ \\ 
\hline 
Teacher VGG-f     &  {\Circle}   &  {\rotatebox[origin=c]{90}{\LEFTcircle}}   & $85.69\%$ \\
Teacher VGG-face  &  {\Circle}   &  {\rotatebox[origin=c]{90}{\LEFTcircle}}   & $88.18\%$ \\
Teacher ResNet-50 &  {\Circle}   &  {\rotatebox[origin=c]{90}{\LEFTcircle}}   & $83.20\%$ \\ 
\hline 
VGG-f             &  {\rotatebox[origin=c]{90}{\RIGHTcircle}} &  {\rotatebox[origin=c]{90}{\RIGHTcircle}}  & $88.70\%$ \\ 
VGG-face          &  {\rotatebox[origin=c]{90}{\RIGHTcircle}} &  {\rotatebox[origin=c]{90}{\RIGHTcircle}}  & $90.62\%$ \\ 
ResNet-50         &  {\rotatebox[origin=c]{90}{\RIGHTcircle}} &  {\rotatebox[origin=c]{90}{\RIGHTcircle}}  & $86.47\%$ \\ 
\hline 
VGG-f             &  {\rotatebox[origin=c]{90}{\LEFTcircle}}  &  {\rotatebox[origin=c]{90}{\LEFTcircle}}   & $88.92\%$ \\  
VGG-face          &  {\rotatebox[origin=c]{90}{\LEFTcircle}}  &  {\rotatebox[origin=c]{90}{\LEFTcircle}}   & $88.26\%$ \\
ResNet-50         &  {\rotatebox[origin=c]{90}{\LEFTcircle}}  &  {\rotatebox[origin=c]{90}{\LEFTcircle}}   & $88.75\%$ \\  
\hline 
VGG-f (standard T-S)   &  {\Circle} + {\rotatebox[origin=c]{90}{\LEFTcircle}}  &  {\rotatebox[origin=c]{90}{\LEFTcircle}}   & $89.13\%$ \\ 
VGG-face (standard T-S) &  {\Circle} + {\rotatebox[origin=c]{90}{\LEFTcircle}}  &  {\rotatebox[origin=c]{90}{\LEFTcircle}}   & $88.45\%$ \\ 
ResNet-50 (standard T-S) &  {\Circle} + {\rotatebox[origin=c]{90}{\LEFTcircle}}  &  {\rotatebox[origin=c]{90}{\LEFTcircle}}   & $89.45\%^\dagger$ \\ 
\hline
VGG-f (triplet loss T-S) &  {\Circle} + {\rotatebox[origin=c]{90}{\LEFTcircle}}  &  {\rotatebox[origin=c]{90}{\LEFTcircle}}   & $89.55\%^\dagger$ \\ 
VGG-face (triplet loss T-S) &  {\Circle} + {\rotatebox[origin=c]{90}{\LEFTcircle}}  &  {\rotatebox[origin=c]{90}{\LEFTcircle}}   & $88.31\%$ \\ 
ResNet-50 (triplet loss T-S) &  {\Circle} + {\rotatebox[origin=c]{90}{\LEFTcircle}}  &  {\rotatebox[origin=c]{90}{\LEFTcircle}}   & $89.19\%$ \\
\hline 
VGG-f (triplet loss + standard T-S) &  {\Circle} + {\rotatebox[origin=c]{90}{\LEFTcircle}} &  {\rotatebox[origin=c]{90}{\LEFTcircle}}   & $89.82\%^\dagger$ \\ 
VGG-face (triplet loss + standard T-S) &  {\Circle} + {\rotatebox[origin=c]{90}{\LEFTcircle}} &  {\rotatebox[origin=c]{90}{\LEFTcircle}}   & $90.35\%^\dagger$ \\ 
ResNet-50 (triplet loss + standard T-S) &  {\Circle} + {\rotatebox[origin=c]{90}{\LEFTcircle}} &  {\rotatebox[origin=c]{90}{\LEFTcircle}}   & $89.63\%^\dagger$ \\ 
\hline 
\end{tabular}
\end{center}
\vspace{-0.3cm}
\end{table*}

\begin{figure*}
\centering
  \includegraphics[width=1.0\textwidth]{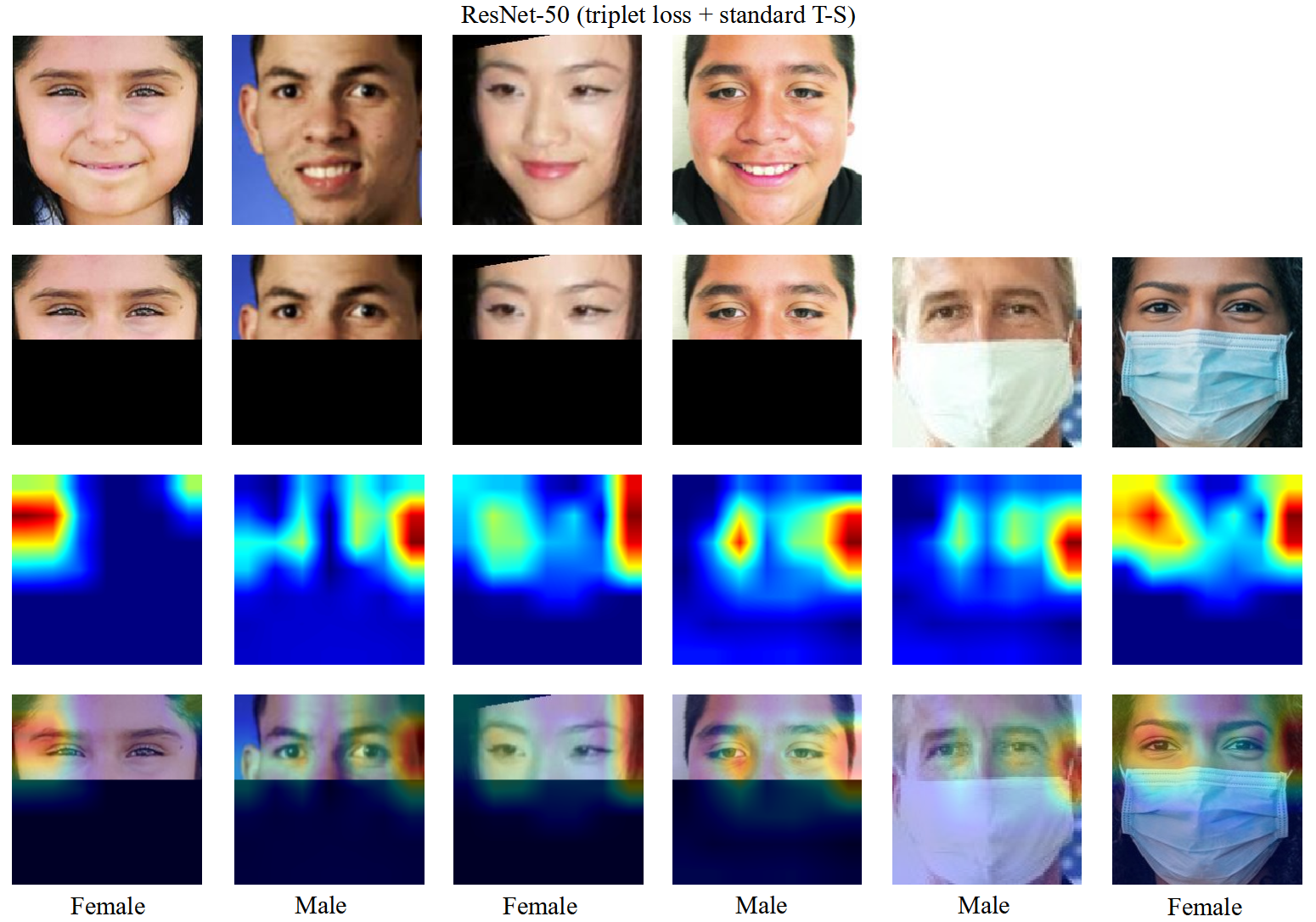}  %
\caption{Fully-visible images ({\Circle}) on top row, upper-half-visible faces ({\protect\rotatebox[origin=c]{90}{\LEFTcircle}}) on second row, Grad-CAM~\cite{Selvaraju-ICCV-2017} explanation masks on third row and upper-half-visible faces with superimposed Grad-CAM masks on bottom row. The predicted gender provided by the distilled ResNet-50 model is shown at the bottom. The first four examples are selected from the UTKFace \cite{Zhang-CVPR-2017} data set. The last two examples are people wearing surgical masks. Best viewed in color.}
\label{fig:gender}       
\end{figure*}

\subsubsection{Gender Recognition Results}
\noindent
{\bf Comparison with the state-of-the-art.}
We present the gender prediction results in Table \ref{tab:gender_resnet_50}. The teacher ResNet-50 trained and evaluated on fully-visible faces (denoted by {\Circle}) reaches an accuracy of $90.88\%$, which is quite close to the top result reported in~\cite{Georgescu-A-2020}. The VGG-f and VGG-face teachers surpass the performance reported in~\cite{Georgescu-A-2020} by at least $2\%$.

\noindent
{\bf Comparison between lower-half and upper-half visible faces.}
When we evaluate the teachers on half-visible faces (denoted by {\rotatebox[origin=c]{90}{\RIGHTcircle}} and {\rotatebox[origin=c]{90}{\LEFTcircle}}), the accuracy rates drop by considerable margins. On the one hand, evaluating the teachers on the lower half ({\rotatebox[origin=c]{90}{\RIGHTcircle}}) of the faces (the upper half being occluded) induces a performance drop between $14\%$ and $19\%$. On the other hand, when we black out the lower half of each face, evaluating the teachers on upper-half-visible faces ({\rotatebox[origin=c]{90}{\LEFTcircle}}), the accuracy decreases by less than $7.68\%$ (the maximum drop being observed for the ResNet-50 teacher), suggesting that it is easier to perform gender recognition on the upper side of the face. Despite the significant performance drop, this represents an encouraging result for the surgical mask scenario.

\begin{table*} 
\caption{Mean absolute error (MAE) values for age estimation on UTKFace~\cite{Zhang-CVPR-2017}, for fully-visible faces (denoted by {\Circle}), lower-half-visible faces (denoted by {\protect\rotatebox[origin=c]{90}{\RIGHTcircle}}) and upper-half-visible faces (denoted by {\protect\rotatebox[origin=c]{90}{\LEFTcircle}}). A state-of-the-art model \cite{Georgescu-A-2020} is included as reference.}
\label{tab:age_resnet_50}
\begin{center}
\begin{tabular}{|l|c|c|c|} 
\hline
\bf Method        & \bf Train faces     & \bf Test faces      & \bf MAE \\
\hline  
\hline
ResNet-50+PyNADA \cite{Georgescu-A-2020}    &  {\Circle}     &  {\Circle}                                 & $5.79$ \\ 
\hline 
Teacher VGG-f     & {\Circle}               &  {\Circle}           & $5.63$ \\ 
Teacher VGG-face  & {\Circle}               &  {\Circle}           & $5.11$ \\ 
Teacher ResNet-50 & {\Circle}               &  {\Circle}           & $5.27$ \\ 
\hline 
Teacher VGG-f     & {\Circle}  &  {\rotatebox[origin=c]{90}{\RIGHTcircle}}  & $11.16$ \\ 
Teacher VGG-face  & {\Circle}  &  {\rotatebox[origin=c]{90}{\RIGHTcircle}}  & $13.08$ \\ 
Teacher ResNet-50 & {\Circle}  &  {\rotatebox[origin=c]{90}{\RIGHTcircle}}  & $14.23$ \\ 
\hline 
Teacher VGG-f     & {\Circle}  &  {\rotatebox[origin=c]{90}{\LEFTcircle}}   & $9.60$ \\
Teacher VGG-face  & {\Circle}  &  {\rotatebox[origin=c]{90}{\LEFTcircle}}   & $10.30$ \\
Teacher ResNet-50 & {\Circle}  &  {\rotatebox[origin=c]{90}{\LEFTcircle}}   & $11.92$ \\
\hline 
VGG-f     &  {\rotatebox[origin=c]{90}{\RIGHTcircle}}  &  {\rotatebox[origin=c]{90}{\RIGHTcircle}}  & $6.80$ \\
VGG-face  &  {\rotatebox[origin=c]{90}{\RIGHTcircle}}  &  {\rotatebox[origin=c]{90}{\RIGHTcircle}}  & $6.15$ \\
ResNet-50 &  {\rotatebox[origin=c]{90}{\RIGHTcircle}}  &  {\rotatebox[origin=c]{90}{\RIGHTcircle}}  & $6.66$ \\
\hline 
VGG-f     &  {\rotatebox[origin=c]{90}{\LEFTcircle}}    &  {\rotatebox[origin=c]{90}{\LEFTcircle}}   & $6.36$ \\  
VGG-face  &  {\rotatebox[origin=c]{90}{\LEFTcircle}}    &  {\rotatebox[origin=c]{90}{\LEFTcircle}}   & $5.53$ \\
ResNet-50 &  {\rotatebox[origin=c]{90}{\LEFTcircle}}    &  {\rotatebox[origin=c]{90}{\LEFTcircle}}   & $6.44$ \\  
\hline
VGG-f (standard T-S)    &   {\Circle} + {\rotatebox[origin=c]{90}{\LEFTcircle}}  &  {\rotatebox[origin=c]{90}{\LEFTcircle}}   & $6.34$ \\ 
VGG-face (standard T-S) &   {\Circle} + {\rotatebox[origin=c]{90}{\LEFTcircle}}  &  {\rotatebox[origin=c]{90}{\LEFTcircle}}   & $5.40$ \\ 
ResNet-50 (standard T-S) &   {\Circle} + {\rotatebox[origin=c]{90}{\LEFTcircle}}  &  {\rotatebox[origin=c]{90}{\LEFTcircle}}   & $6.35$ \\ 
\hline
VGG-f (triplet loss T-S) &   {\Circle} + {\rotatebox[origin=c]{90}{\LEFTcircle}}   &  {\rotatebox[origin=c]{90}{\LEFTcircle}}   & $6.34$ \\ 
VGG-face (triplet loss T-S) &   {\Circle} + {\rotatebox[origin=c]{90}{\LEFTcircle}}   &  {\rotatebox[origin=c]{90}{\LEFTcircle}}   & $5.42$ \\ 
ResNet-50 (triplet loss T-S) &   {\Circle} + {\rotatebox[origin=c]{90}{\LEFTcircle}}   &  {\rotatebox[origin=c]{90}{\LEFTcircle}}   & $6.34$ \\ 
\hline
VGG-f (triplet loss + standard T-S)   &  {\Circle} +  {\rotatebox[origin=c]{90}{\LEFTcircle}} &  {\rotatebox[origin=c]{90}{\LEFTcircle}}   & $6.22$ \\ 
VGG-face (triplet loss + standard T-S)   &  {\Circle} +  {\rotatebox[origin=c]{90}{\LEFTcircle}} &  {\rotatebox[origin=c]{90}{\LEFTcircle}}   & $5.40$ \\ 
ResNet-50 (triplet loss + standard T-S)   &  {\Circle} +  {\rotatebox[origin=c]{90}{\LEFTcircle}} &  {\rotatebox[origin=c]{90}{\LEFTcircle}}   & $6.33$ \\ 
\hline 
\end{tabular}
\end{center}
\end{table*}

When we train the students on half-visible faces and evaluate them in the same setting, the accuracy rates improve. The ResNet-50 student trained and evaluated on the lower-half-visible faces reaches an accuracy of $86.47\%$. The other ResNet-50 student, the one trained and evaluated on upper-half-visible faces, yields an accuracy of $88.75\%$. Both students obtain better performance compared with the teacher trained on fully-visible faces, when the evaluation is conducted on half-visible faces. 

\noindent
{\bf Comparison with the baseline.}
Further, we observe that, with respect to the baseline students, both teacher-student strategies attain superior performance.
When we fine-tune the students using the standard teacher-student strategy, we obtain improvements ranging between $0.19\%$ and $0.70\%$, reaching a top accuracy rate of $89.45\%$ with the ResNet-50 model. The privileged information received from the teacher helps the student to outperform its ablated version trained with the standard loss (binary cross-entropy between the predicted labels and the ground-truth labels). The VGG-f student fine-tuned using the teacher-student training strategy based on triplet loss obtains an accuracy of $89.55\%$, surpassing its ablated version by $0.63\%$. Concatenating the VGG-face embeddings of the two knowledge distillation strategies provides an accuracy of $90.35\%$. We emphasize that the performance of the ensemble formed by the two ResNet-50 students evaluated on upper-half-visible faces ({\rotatebox[origin=c]{90}{\LEFTcircle}}) is only $1.25\%$ below the ResNet-50 teacher trained and evaluated on fully-visible faces ({\Circle}). We thus conclude that putting on a mask does not represent a significant problem for gender recognition when both fine-tuning and knowledge distillation are employed.

\noindent
{\bf Statistical significance testing.}
Furthermore, we also performed statistical significance testing to compare our distilled models with the ablated version of the student (the version before undergoing distillation). For all network types (VGG-f, VGG-face and ResNet-50), the results obtained by concatenating the two strategies and those obtained by the model fine-tuned using the standard teacher-student paradigm are indeed statistically significant at a significance level of $0.05$.

\noindent
{\bf Grad-CAM visualizations.}
To further investigate how our models make decisions, we employ Grad-CAM~\cite{Selvaraju-ICCV-2017} to visualize what parts of the image are important in the gender recognition process. A set of representative Grad-CAM~\cite{Selvaraju-ICCV-2017} visualizations are shown in Figure~\ref{fig:gender}. We observe that our model tends to concentrate on the upper half of the face, especially on the eye region. We notice that the focus area usually extends until it covers some hair. Thus, it is likely that our model considers the shape of the eyes and the length of the hair as discriminative features for gender prediction. Additionally, we underline that the predicted classes for the samples presented in Figure~\ref{fig:gender} are all correct.

\subsubsection{Age Estimation Results}
   
\begin{figure*}
\centering
  \includegraphics[width=1.0\textwidth]{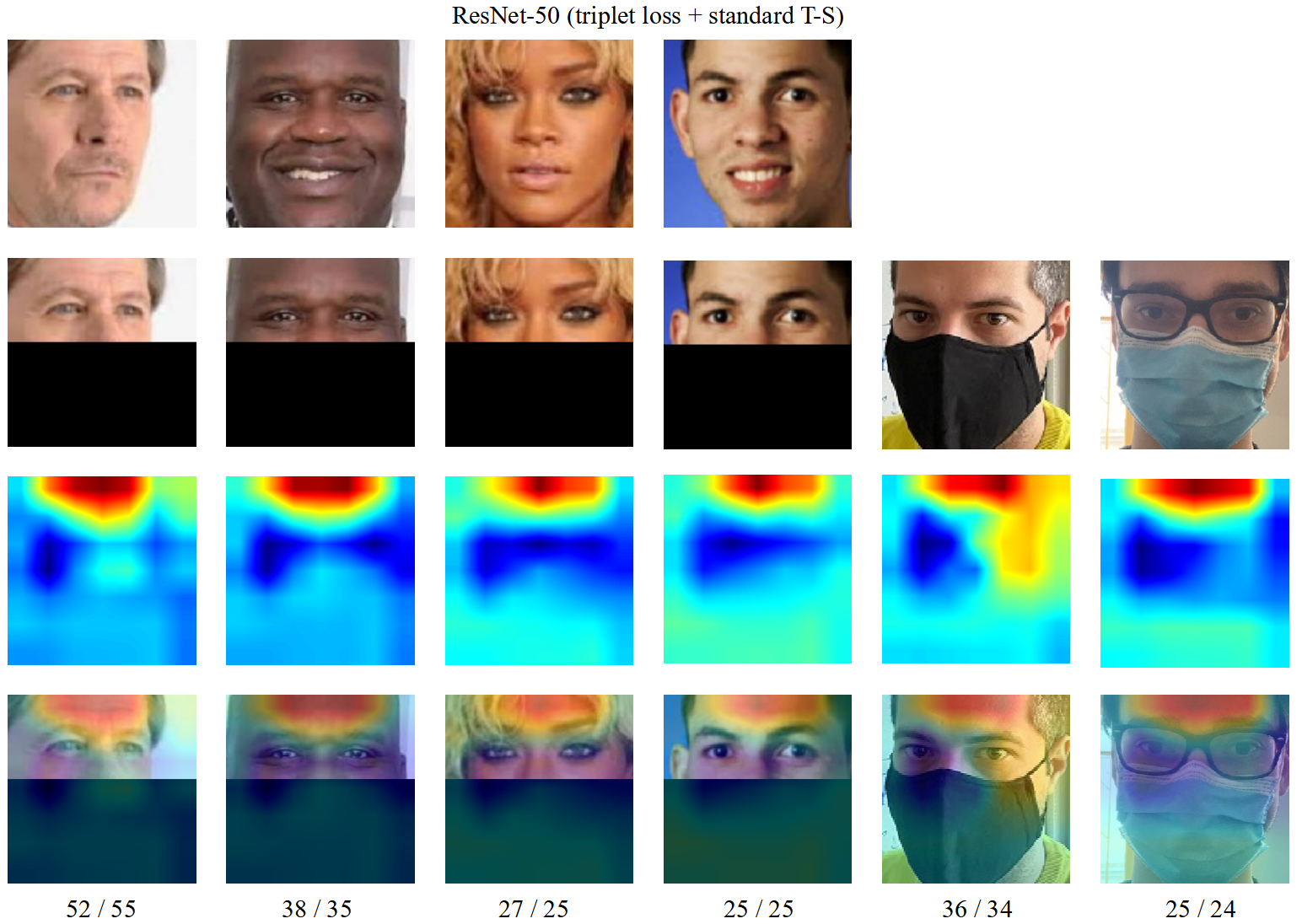}  %
\caption{Fully-visible images ({\Circle}) on top row, upper-half-visible faces ({\protect\rotatebox[origin=c]{90}{\LEFTcircle}}) on second row, Grad-CAM~\cite{Selvaraju-ICCV-2017} explanation masks on third row and upper-half-visible faces with superimposed Grad-CAM masks on bottom row. The estimated age provided by the distilled ResNet-50 model is the first number shown at the bottom. The second number is the ground-truth age. The first four examples are selected from the UTKFace~\cite{Zhang-CVPR-2017} data set, while the last two examples are people wearing masks. Best viewed in color.}
\label{fig:age}       
\end{figure*}

\noindent
{\bf Comparison with the state-of-the-art.}
In Table \ref{tab:age_resnet_50}, we present the results for the age estimation task. The teacher ResNet-50 trained and evaluated on fully-visible faces (denoted by {\Circle}) obtains an error of $5.27$ years, surpassing the reference model proposed in \cite{Georgescu-A-2020}.

\noindent
{\bf Comparison between lower-half and upper-half visible faces.}
When we evaluate the teacher models on the half-visible faces, the error increases by considerable margins. Indeed, evaluating the teacher models on lower-half-visible faces (denoted by {\rotatebox[origin=c]{90}{\RIGHTcircle}}) increases the error up to $14.23$ years. Similarly, when we evaluate them on upper-half-visible faces (denoted by {\rotatebox[origin=c]{90}{\LEFTcircle}}), the MAE grows up to $11.92$ years.

Training the neural networks on half-visible faces and evaluating them in the same manner reduces the error by large margins. More precisely, the MAE goes down to $6.15$ years on lower-half-visible faces and $5.53$ years on upper-half-visible faces. Based on the results discussed above, we conclude that the upper half of the face is more informative for age estimation than the lower half.

\noindent
{\bf Comparison with the baseline.} 
When fine-tuning the student models with the standard teacher-student training strategy, the error decreases from $6.44$ to $6.35$ years for the ResNet-50 model, and from $5.53$ to $5.40$ years for the VGG-face model. Fine-tuning the ResNet-50 student using our teacher-student training paradigm based on triplet loss reduces the error to $6.34$ years. By concatenating the embeddings of the two VGG-f students, we obtain a MAE of $6.22$, this being our best improvement over the ablated version which scores a MAE of $6.36$ years. In the end, the maximum difference between the teacher model trained and evaluated on fully-visible faces and the student evaluated on lower-half-visible faces is $1.07$ years. Hence, we conclude that the negative performance impact on age estimation generated by putting on a mask can be reduced by employing both fine-tuning and knowledge distillation.

\noindent
{\bf Grad-CAM visualizations.}
To explain how our distilled ResNet-50 models make decisions, we employ Grad-CAM \cite{Selvaraju-ICCV-2017} to observe what parts of the image are seen as important when estimating the age. We provide some relevant Grad-CAM \cite{Selvaraju-ICCV-2017} visualizations in Figure~\ref{fig:age}. First, we observe that our student model tends to focus on the upper half of the face, especially on the forehead region. We conjecture that our model views the wrinkles formed on the forehead as a discriminative feature for age estimation. On the bottom of the figure, we show the predicted age (first number) and the ground-truth age (second number) for each image. For the selected samples, the maximum difference between the predicted and the actual age is $3$ years. We consider these predictions as fairly accurate.

\begin{table}
\setlength\tabcolsep{2.0pt}
\caption{The accuracy rates of SVMs trained on embeddings extracted from students based on standard teacher-student (TS) or triplet loss (TL) strategies. These models are compared with SVMs trained on concatenated embeddings as well as the students providing the embeddings. Results are reported for two tasks: facial expression recognition (on FER+ and AffectNet) and gender prediction (on UTKFace).}
\label{tab:ablation_svm}
\begin{center} 
\begin{tabular}{ |l|c|c|c|c| }  
\hline
\bf Network                     & \bf Method       & \bf FER+  &  \bf AffectNet  &  \bf UTKFace \\ 
\hline
\hline
\multirow{5}{*}{VGG-f}      & TS  & $80.17\%$  & $48.75\%$  & $89.13\%$ \\
                            & TL  & $80.05\%$  & $48.13\%$  & $89.55\%$ \\
                            & TS+SVM        & $80.39\%$ &  $48.52\%$   &  $89.04\%$ \\ 
                            &TL+SVM        & $79.06\%$ &  $47.01\%$   &  $89.70\%$ \\ 
                            &TS+TL+SVM     & $81.09\%$ &  $48.70\%$   &  $89.82\%$\\       
\hline 
\multirow{5}{*}{VGG-face}    & TS  & $82.37\%$  & $49.75\%$  & $88.45\%$ \\
                             & TL  & $82.57\%$  & $49.71\%$  & $88.31\%$ \\
                             &TS+SVM      & $82.34\%$  &  $48.89\%$    &  $90.35\%$ \\    
                              &TL+SVM      & $82.37\%$  &  $49.90\%$    &  $90.27\%$ \\
                              &TS+TL+SVM   & $82.75\%$  &  $50.09\%$    &  $90.35\%$\\    
\hline
\end{tabular}
\end{center}
\end{table}

\subsection{Ablation Study Regarding Neural Embeddings}
\label{sec_ablation}

To understand which part of the ensemble of distilled models brings a larger improvement, i.e.~the concatenation of embeddings or the SVM model, we conduct an ablation study by training an SVM on top of each type of embedding extracted from the distilled students (without concatenating the embeddings). We present the corresponding results in Table~\ref{tab:ablation_svm}. The experiments are performed for the VGG-f and VGG-face models on the FER+, AffectNet and UTKFace data sets. For the facial expression recognition data sets (FER+, AffectNet), we select the distilled models trained on lower-half-visible faces, while for the gender prediction data set (UTKFace), we select the distilled models trained on upper-half-visible faces. For the VGG-f model, we observe that the largest improvement is brought by the concatenation of embeddings, not by the SVM model. However, on the UTKFace data set, the SVM model trained on embeddings from the VGG-face based on triplet loss performs on par with the SVM ensemble of distilled models. In this particular case, it seems that the SVM itself makes the difference. Nevertheless, none of the SVMs trained on individual embeddings is able to surpass the performance of the SVM ensemble. We thus conclude that concatenating the embeddings is useful.

\section{Conclusion}
\label{sec_conclusion}
 
In this paper, we presented two teacher-student methods to improve the performance of neural network models evaluated in scenarios with strong face occlusion. We demonstrate that our methods generalize across different classification and regression tasks (facial expression recognition, age estimation, gender prediction) and neural architecture types (VGG-face, VGG-f, ResNet-50). To the best of our knowledge, we are the first to study teacher-student strategies to learn privileged information aiming to cope with strong occlusions in images. We also proposed a novel teacher-student method based on triplet loss. The empirical results suggest that our knowledge distillation methods obtain superior performance over the baselines. On facial expression recognition, our ensemble of distilled models is only $2.24\%$ below the performance of VGG-13~\cite{Barsoum-ICMI-2016}, when the former method is evaluated on occluded faces and the latter method is evaluated on fully-visible faces. Similarly, on gender recognition, our ensemble of distilled ResNet-50 models is only $1.25\%$ below the corresponding teacher. On age estimation, the difference between our ensemble of distilled ResNet-50 models applied on occluded faces and the teacher applied on fully-visible faces is $1.06$ years. In conclusion, we believe that the performance levels of our distilled models are sufficiently high to be used in real-life scenarios, such as changing the environment in VR applications based on user's emotions and estimating the age and gender of people wearing surgical masks in supermarkets or retail stores.

In future work, we aim to study if models trained under occlusion can be used to boost the performance of teachers on fully-visible faces. Our future goal is to find an effective way to combine students specialized on specific parts of the face with teachers looking at the entire face. We believe that this line of research can lead to significant performance gains.

\begin{acknowledgements}
We thank reviewers for their valuable feedback which led to important improvements of the article.
\end{acknowledgements}

%
%

\bibliographystyle{spmpsci}      

%
%

\bibliography{references}

\end{document}